\documentclass[runningheads]{llncs}

\usepackage{graphicx}
\usepackage{comment}
\usepackage{amsmath,amssymb} %

\usepackage{booktabs}
\usepackage{wrapfig}
\usepackage{tabularx}
\usepackage{longtable}
\usepackage{epstopdf}
\usepackage{caption}
\usepackage[permil]{overpic}
\usepackage{subcaption}
\usepackage{arydshln}
\usepackage{xcolor}
\usepackage{colortbl}

\usepackage{microtype} 

\usepackage{pbox}
\usepackage{pifont}
\usepackage{array} %

\usepackage{algorithm}
\usepackage{algpseudocode}
\algnewcommand{\LineComment}[1]{\(\triangleright\)\color{white!15!black}{ #1}\color{black}}
\algrenewcommand\algorithmicindent{1em}%

\usepackage{etoolbox}

\usepackage[pagebackref=false,breaklinks=true,letterpaper=true,colorlinks,bookmarks=false]{hyperref}

\newcommand{\temppara}[1]{}

\renewcommand{\vec}[1]{\boldsymbol{\mathbf{#1}}}

\def\bx{\vec{x}}
\def\by{\vec{y}}
\def\bc{\vec{c}}

\def\bp{\vec{p}}

\newcommand{\cnn}[0]{\vec{f}}
\def\btheta{\vec{\theta}}
\def\deriv{\mathrm{d}}
\def\CE{\mathcal{L}_{\mathrm{CE}}}

\def\GCE{\mathcal{L}_{\mathrm{GCE}}}
\def\MAS{\mathcal{L}_{\mathrm{F}}}
\def\dataset{\mathcal{D}}
\def\Ladapt{\mathcal{L}_{\mathrm{ADAPT}}}
\def\indicator{\vec{1}}

\newcommand{\subsec}[1]{\subsection{#1}}
\newcommand{\para}[1]{\par\noindent\textbf{#1}}
\newcolumntype{H}{>{\setbox0=\hbox\bgroup}c<{\egroup}@{}}

\newcolumntype{C}[1]{>{\centering\arraybackslash\hspace{0pt}}p{#1}}

\providetoggle{showcomments}									%
\settoggle{showcomments}{false} 								%

\providetoggle{appendsupplementary}						 	%
\settoggle{appendsupplementary}{true} 							%

\iftoggle{showcomments}{%
\newcommand{\todo}[1]{\textcolor{blue}{\textbf{TODO:} #1}}
\newcommand{\changed}[1]{\textcolor{blue}{#1}}

\newcommand{\michael}[1]{\textcolor{orange}{\textbf{Michael:} #1}}
\newcommand{\dora}[1]{\textcolor{violet}{\textbf{Dora:} #1}}
\newcommand{\vitto}[1]{\textcolor{red}{\textbf{Vitto:} #1}}
\newcommand{\jasper}[1]{\textcolor{magenta}{\textbf{Jasper:} #1}}
\newcommand{\att}[1]{\textcolor{red}{#1}}
}{%
\newcommand{\changed}[1]{{#1}}
\newcommand{\todo}[1]{}

\newcommand{\michael}[1]{}
\newcommand{\dora}[1]{}
\newcommand{\vitto}[1]{}
\newcommand{\jasper}[1]{}
\newcommand{\att}[1]{\textcolor{black}{#1}}}

\newcommand{\datasetSGD}{image sequence adaptation}
\newcommand{\imageSGD}{single image adaptation}
\newcommand{\basemodel}{initial model}
\newcommand{\fixedmodel}{frozen}
\newcommand{\fixedmodellong}{frozen model}

\newcommand{\classspecialisation}{class specialization}

\newcommand{\ade}{ADE20k}
\newcommand{\pascal}{PASCAL VOC12}
\newcommand{\davis}{DAVIS16}

\newcommand{\coco}{COCO}

\newcommand{\cocounseenlarge}{COCO unseen 6k}
\newcommand{\rooftop}{Rooftop Aerial}
\newcommand{\medical}{DRIONS-DB}
\newcommand{\grabcut}{GrabCut}
\newcommand{\berkeley}{Berkeley}
\newcommand{\youtubeVOS}{YouTube\,-VOS}

\newcommand{\clicksAtIOU}{clicks@q\%}
\newcommand{\clicksAt}[1]{clicks@#1\%}
\newcommand{\iouAtK}{IoU@k}
\newcommand{\iouAt}[1]{IoU@#1}
\newcommand{\iou}{IoU}
\newcommand{\gt}{ground-truth}

\newcommand{\imageadaptationshort}{IA}

\newcommand{\sequenceadaptationshort}{SA}

\newcommand{\combinedshort}{IA+SA}

\newcommand{\eg}{\textit{e.g.~}}
\newcommand{\ie}{\textit{i.e.~}}

\newcommand{\etal}{\textit{et al.}}
\newcommand{\vs}{\textit{vs.~}}
\newcommand{\uc}{\expandafter\MakeUppercase}

\makeatletter
\newcommand{\printfnsymbol}[1]{%
  \textsuperscript{\@fnsymbol{#1}}%
}
\def\@fnsymbol#1{\ensuremath{\ifcase#1\or *\or  \ddagger\or
   \mathsection\or \mathparagraph\or \|\or **\or \dagger\dagger
   \or \ddagger\ddagger \else\@ctrerr\fi}}
\makeatother

\begin{document}
\frenchspacing %
\pagestyle{headings}
\mainmatter

\title{Continuous Adaptation for Interactive Object Segmentation by Learning from Corrections}
\titlerunning{Adaptive Interactive Segmentation by Learning from Corrections}
\authorrunning{Kontogianni\printfnsymbol{1} \and Gygli\printfnsymbol{1} \and Uijlings \and Ferrari}
\author{Theodora Kontogianni\thanks{Equal contribution}\thanks{Work done while interning at Google.} \inst{2} \and
Michael Gygli\printfnsymbol{1} \inst{1} \and \\
Jasper Uijlings \inst{1} \and
Vittorio Ferrari \inst{1}}
\institute{%
Google Research, Zurich\and
RWTH Aachen University, Germany}
\maketitle

\def\httilde{\mbox{\tt\raisebox{-.5ex}{\symbol{126}}}}

\begin{abstract}

In interactive object segmentation a user collaborates with a computer vision model to segment an object.
Recent works employ convolutional neural networks for this task: Given an image and a set of corrections made by the user as input,
they output a segmentation mask.
These approaches achieve strong performance by training on large datasets but they keep the model parameters unchanged at test time.
Instead, we recognize that user corrections can serve as sparse training examples and we propose a method that capitalizes on that idea to update the model parameters on-the-fly to the data at hand.
Our approach enables the adaptation to a particular object and its background, to distributions shifts in a test set, to specific object classes, and even to large domain changes, where the imaging modality changes between training and testing.
We perform extensive experiments on 8 diverse datasets and show:
Compared to a model with \fixedmodel{} parameters, our method reduces the required corrections 
(i)~by \att{9\%-30\%} when distribution shifts are small between training and testing;
(ii)~by \att{12\%-44\%} when specializing to a specific class;
(iii)~and by \att{60\%} and \att{77\%} when we completely change domain between training and testing.

\end{abstract}

\section{Introduction}
\label{sec:introduction}
In interactive object segmentation a human collaborates with a computer vision model to segment an object of interest~\cite{boykov01iccv,Rother04-tdfixed,xu16cvpr,benenson19cvpr}.
The process iteratively alternates between the user providing corrections on the current segmentation and
the model refining the segmentation based on these corrections.
The objective of the model is to infer an accurate segmentation mask from as few user corrections as possible (typically point clicks~\cite{bearman16eccv,chen18aaai} or strokes~\cite{Rother04-tdfixed,gulshan10cvpr} on mislabeled pixels).
This enables fast and accurate object segmentation, which is indispensable for image editing~\cite{photoshop18}
and collecting ground-truth segmentation masks at scale~\cite{benenson19cvpr}.

\begin{figure}
\hskip 0.5cm
\centering
\begin{overpic}[trim={1.5cm 0.5cm 0 0cm},clip,width=0.85\linewidth]{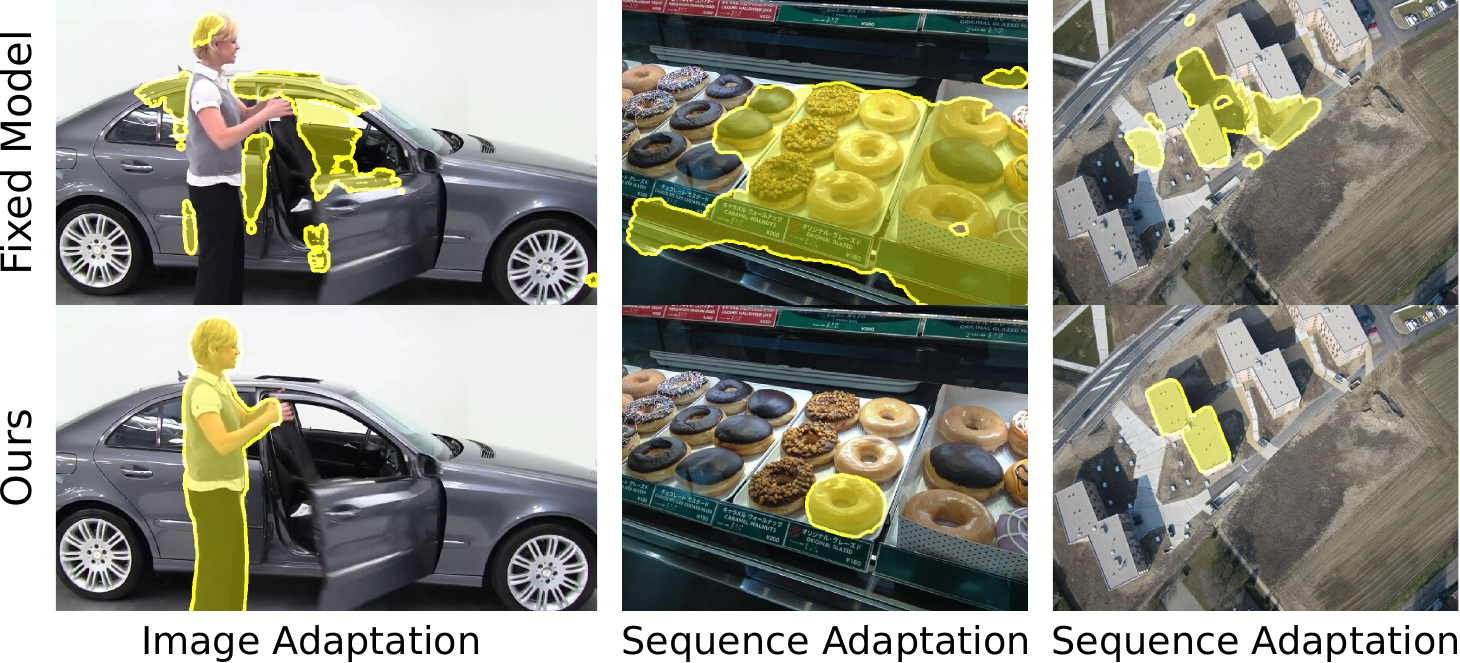}
\put(-35,350){\makebox(0,0){\rotatebox{90}{\uc \fixedmodellong{}~\cite{mahadevan18bmvc}}}}
\put(-35,100){\makebox(0,0){\rotatebox{90}{\textbf{Ours}}}}
\end{overpic}
\caption{
\textbf{Example results for a \fixedmodel{} model (top) and our adaptive methods (bottom).}
A \fixedmodel{} model performs poorly when foreground and background share similar appearance (left), when it is used to segment new object classes absent in the training set (center, donut class), or when the model is tested on a different image domain\changed{~(aerial)} than it is trained on\changed{~(consumer)} (right).
By using corrections to adapt the model parameters to a specific test image, or to the test image sequence, our method substantially improves segmentation quality.
The input is four corrections~in~all~cases~shown.
}
\label{fig:teaser}
\end{figure}

Current state-of-the-art methods train a convolutional neural network (CNN) which takes \changed{an} image and user corrections as input and predicts
a foreground~/~background segmentation \cite{xu16cvpr,liew17iccv,benard17arxiv,mahadevan18bmvc,li18cvpr,benenson19cvpr,jang19cvpr}.
At test time, the model parameters are frozen
and corrections are \changed{only} used as additional input to guide the model predictions.
But in fact, user corrections directly specify the ground-truth labelling of the corrected pixels.
In this paper we capitalize on this observation: we treat user corrections as
training examples to adapt our model on-the-fly.
We use these user corrections in two ways:\changed{
(1) in \emph{\imageSGD{}} we iteratively adapt
model parameters to \changed{one} specific object in an image, given the corrections produced while segmenting that object;
(2) in \emph{\datasetSGD} we adapt model parameters to a sequence of images with an online method, given the set of corrections produced on these images.
}
Each of these leads to distinct advantages over using a \fixedmodel{} model:

During {\em \imageSGD} our model learns the specific appearance of the current object instance and the surrounding background.
This allows the model to adapt even to subtle differences between foreground and background for that specific example. This is necessary when the object to be segmented has similar color to the background (Fig.~\ref{fig:teaser}, 1st column), has blurry object boundaries, or low contrast.
In addition, a \fixedmodel{} model can sometimes ignore the user corrections and overrule them in its next prediction. We avoid this undesired behavior by updating the model parameters until its predictions respect the~user~corrections.

During {\em \datasetSGD} we continuously adapt the model to a \changed{sequence of segmentation tasks.}
Through this, the model parameters are optimized to the image and class distribution in these tasks, which may consist of
 different types of images or a set of new classes which are unseen during training.
An important case of this is specializing the model for segmenting objects of a single class. This is useful for collecting many examples in high-precision domains, such as \emph{pedestrians} for self-driving car applications. Fig.~\ref{fig:teaser}, middle column shows an example of specializing to the single, unseen class \emph{donut}.
Furthermore, an important property of image sequence adaptation is that it enables us to handle large domain changes, where the imaging modality changes dramatically between training and testing. We demonstrate this by training on consumer photos while testing on medical and aerial images (Fig.~\ref{fig:teaser}, right column).

Naturally, \imageSGD{} and \datasetSGD{} can be used jointly, leading to a method that combines their advantages.

In summary: Our innovative idea of treating user corrections as training examples allows to update the parameters of an interactive segmentation model at {\em test time}. To update the parameters we propose a practical online adaptation method. Our method operates on sparse corrections, balances adaptation \vs retaining old knowledge and can be applied to any CNN-based interactive segmentation model.
We perform extensive experiments on 8 diverse datasets and show:
Compared to a model with \fixedmodel{} parameters, our method reduces the required corrections 
(i) by \att{9\%-30\%} when distribution shifts are small between training and testing;
(ii) by \att{12\%-44\%} when specializing to a specific class;
(iii) and by \att{60\%} and \att{77\%} when we completely change domain between training and testing.
(iv) Finally,  we evaluate on four standard datasets where distribution shifts between training and testing are minimal.
Nevertheless, our method did set a new state-of-the-art on all of them, when it was initially released~\cite{kontogianni19arxiv}.
\michael{Dora and Vitto will decide it if we leave this as is, just say we are competitive or remove the STOA point entirely (removing is Michaels least favorite option).}

\section{Related Work}
\label{sec:related_work}
\para{Interactive Object Segmentation.}
Traditional methods approach interactive segmentation via energy minimization on a graph defined over pixels~\cite{boykov01iccv,Rother04-tdfixed,bai09ijcv,gulshan10cvpr,price10cvpr}.
User inputs are used to create an image-specific appearance model based on low-level features (\eg color), which is then used to predict foreground and background probabilities. A pairwise smoothness term between neighboring pixels encourages regular segmentation outputs. 
Hence these classical methods are based on a weak appearance model which is specialized to one specific image.

Recent methods rely on Convolutional Neural Networks (CNNs) to interactively produce a segmentation mask~\cite{xu16cvpr,liew17iccv,benard17arxiv,mahadevan18bmvc,li18cvpr,chen18aaai,hu19nn,jang19cvpr,agustsson19cvpr}.
These methods take the image and user corrections (transformed into a guidance map) as input and map them to foreground and background probabilities.
This
mapping is optimized over a training dataset and remains frozen at test time.
Hence these models have a strong appearance model but it is not optimized for the test image or dataset at hand.

Our method combines the advantages of traditional and recent approaches: We use a CNN to learn a strong initial appearance model from a training set. During segmentation of a new test image, we adapt the model to it.
\changed{It thus learns an appearance model specifically for that image.
Furthermore, we also continuously adapt the model to the new image and class distribution of the test set, which may be significantly different from the one the model is originally~trained~on.}

\para{Gradient Descent at test time.}
Several methods iteratively minimize a loss at test time.
The concurrent work of~\cite{sun19arxiv} uses self-supervision to adapt the feature extractor of a multi-tasking model to the test distribution. Instead, we directly adapt the full model by minimizing the task loss.
Others iteratively update the inputs of a model~\cite{gatys16cvpr,gygli17icml,jang19cvpr},
~\eg for style transfer~\cite{gatys16cvpr}.
In the domain of interactive segmentation,~\cite{jang19cvpr} updates the guidance map which encodes the user corrections and is input to the model. \cite{sofiiuk20cvpr} made this idea more computationally efficient by updating intermediate feature activations, rather than the guidance maps.
Instead, our method updates the model parameters, making it more general and allowing it to adapt to individual images as well as sequences.

\para{In-domain Fine-Tuning.}
In other applications it is common practice to fine-tune on in-domain data when transferring a model to a new domain~\cite{caelles17cvpr,papadopoulos16cvpr,voigtlaender17,zhouz17cvpr}.
For example, when supervision for the first frame of a test video is available \cite{perazzi16cvpr,voigtlaender17,caelles17cvpr}, or after annotating a subset of an image dataset~\cite{papadopoulos16cvpr,zhouz17cvpr}.
In interactive segmentation the only existing attempt is~\cite{acuna18cvpr}, which performs polygon annotation~\cite{castrejon17cvpr,acuna18cvpr,ling2019cvpr}. However,
it does not consider adapting to a particular image;
their process to fine-tune on a dataset involves 3 different models, so they do it only a few times per dataset;
they cannot directly train on user corrections, only on complete masks from previous images;
finally, they require a bounding box on the object as input. 

\para{Few-shot and Continual Learning.}
Our method automatically adapts to distribution shifts and domain changes.
It performs domain adaptation from limited supervision, similar to few-shot learning~\cite{ravi16iclr,finn17icml,snell17nips,qi18cvpr}.
It also relates to continual learning~\cite{rebuffi17cvpr,farquhar18arxiv}, except that the output label space of the classifier is fixed.
As in other works, our method needs to balance between preserving existing knowledge and adapting to  new data.
This is often done by fine-tuning on new tasks while discouraging large changes in the network parameters, either by penalizing changes to important parameters~\cite{kirkpatrick17pnas,zenke17icml,aljundi18eccv,aljundi19cvpr} or \hspace{-5pt} changing predictions of the model on old tasks~\cite{li2017learning,shmelkov17iccv,michieli19iccvw}.
Alternatively, some training data of the old task is kept and the model is trained on a mixture of the old and new task data~\cite{rebuffi17cvpr,belouadah19iccv}.

\section{Method}
\label{sec:method}
We adopt a typical interactive object segmentation process~\cite{boykov01iccv,xu16cvpr,mahadevan18bmvc,li18cvpr,benenson19cvpr,jang19cvpr}:
the model is given an image and makes an initial foreground\,/\,background prediction for every pixel.
The prediction is then overlaid on the image and presented to the user, who is asked to make a correction.
The user clicks on a single pixel to mark that it was incorrectly predicted to be foreground instead of background or vice versa. The model then updates the predicted segmentation based on all corrections received so far.
This process iterates until the segmentation reaches a desired quality level.

We start by describing the model we build on (Sec.~\ref{sec:base_network}).
Then, we describe our core contribution: treating user corrections as training examples to adapt the model on-the-fly at test-time (Sec.~\ref{sec:adaptation}).
Lastly, we describe how we simulate user corrections to train and test our method (Sec.~\ref{sec:simulating_users}).

\subsec{Interactive Segmentation Model}
\label{sec:base_network}
As the basis of our approach, we use a strong re-implementation of~\cite{mahadevan18bmvc},
an interactive segmentation model based on a convolutional neural network.
The model takes an RGB image and the user corrections as input and produces a segmentation mask.
As in~\cite{benenson19cvpr} we encode the position of user corrections by placing binary disks into a \textit{guidance map}. This map has the same resolution as the image and consists of two channels (one channel for foreground and one for background corrections).
The guidance map is concatenated with the RGB image to form a 5-channel map $\bx$ which is provided as input to the network.

We use DeepLabV3+~\cite{chen18eccv} as our network architecture, which has demonstrated good performance on semantic segmentation.
However, we note that our method does not depend on a specific architecture and can be used with others as well.

For training the model we need a training dataset $\dataset$ with ground-truth object segmentations, as well as user corrections which we simulate as in~\cite{mahadevan18bmvc} (Sec.~\ref{sec:simulating_users}).
We train the model using the cross-entropy loss over all pixels in an image:
\begin{equation}
\begin{aligned}
  \CE(\bx, \by; \btheta) =  \frac{1}{|\by|}  \{-\by \log \cnn(\bx; \btheta) 
  - (1 - \by) \log (1 - \cnn(\bx; \btheta))\}
\end{aligned}
\label{eq:ce-loss}
\end{equation}
where $\bx$ is the 5-channel input defined above (image plus guidance maps), $\by \in \{0,1\}^{H\times W}$ are the pixel labels of the ground-truth object segmentations, and $\cnn(\bx; \btheta)$ represents the mapping of the convolutional network parameterized by $\btheta$.
$|\cdot|$ denotes the $l_1$ norm.

We produce the initial parameters $\btheta^*$ of the segmentation model by minimizing $\sum_{(\bx_i,\by_i) \in \dataset}   \CE(\bx_i, \by_i; \btheta)$ over the training set using stochastic gradient descent.

\subsection{Learning from Corrections at Test-Time}
\label{sec:adaptation}
\begin{figure}[t]
\centering
\includegraphics[width=1\linewidth]{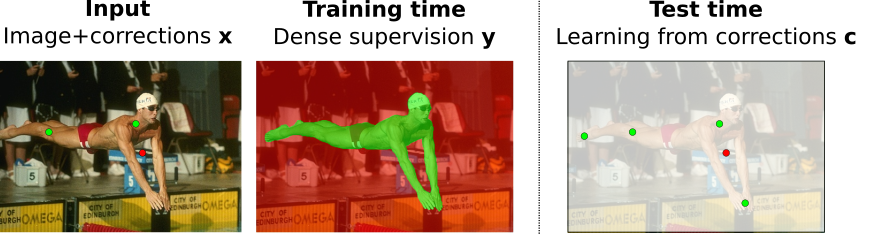}
\caption{\textbf{Corrections as training examples.}
For learning the initial model parameters, full supervision is available, allowing to compute a loss over all the pixels in the image.
At test time, the user provides sparse supervision in the form of corrections. %
We use these to adapt the model parameters.}

\label{fig:adaptation_task}
\end{figure}
Previous interactive object segmentation methods do not treat corrections as training examples.
Thus, the model parameters remain unchanged/\fixedmodel{} at test time \cite{xu16cvpr,benard17arxiv,mahadevan18bmvc,li18cvpr,benenson19cvpr,jang19cvpr} and corrections are only used as inputs to guide the predictions.
Instead, we treat corrections as ground-truth labels to adapt the model at test time.
We achieve this by minimizing the generalized cross-entropy
loss over the corrected pixels:
\begin{align}
\GCE(\bx, \bc; \btheta) = \frac{\indicator[{\scriptstyle \bc \neq -1}]^T}{|\indicator[{\scriptstyle \bc \neq -1}]|} \bigg\{-\bc \log \cnn(\bx; \btheta) 
- (1 - \bc) \log \left(1 - \cnn\left(\bx; \btheta\right)\right) \bigg\}
\label{eq:ce-generalized-loss}
\end{align}
where $\indicator$ is an indicator function and $\bc$ is a vector of values $\{1,0,-1\}$, indicating what pixels were corrected to what label.
Pixels that were corrected to be positive are set to $1$ and negative pixels to $0$. The remaining ones  are set to $-1$, so that they are ignored in the loss.
As there are very few corrections available at test time, this loss is computed over a sparse set of pixels.
This is in contrast to the initial training which had supervision at every pixel
(Sec.~\ref{sec:base_network}).
We illustrate the contrast between the two forms of supervision in Fig.~\ref{fig:adaptation_task}.

\para{Dealing with label sparsity.}
In practice, corrections $\bc$ are extremely sparse and consist of just a handful of scattered points (Fig.~\ref{fig:distribution_adaptations}).
Hence, they offer limited information on the spatial extent of objects and special care needs to be taken to make this form of supervision useful in practice.
As our model is initially trained with full supervision, it has learned strong shape priors. Thus, we propose two auxiliary losses to prevent forgetting these priors as the model is adapted.
First, we regularize the model by treating the initial mask prediction $\bp$ as \gt{} and making it a target in the cross-entropy loss,~\ie $\CE(\bx, \bp; \btheta)$.
This prevents the model from focusing only on the user corrections while forgetting the initially good predictions on pixels for which no corrections were given.

\changed{
Second, inspired by methods for class-incremental learning~\cite{kirkpatrick17pnas,zenke17icml,aljundi18eccv}, we minimize unnecessary changes to the network parameters to prevent it from forgetting crucial patterns learned on the initial training set. Specifically, we add a cost for changing important network parameters:}
\begin{equation}
\begin{aligned}
\MAS(\btheta) = \vec{\Omega}^T\left(\vec{\theta} - \vec{\theta^*}\right)^{\odot 2}
\end{aligned}
\label{eq:mas-loss}
\end{equation}
where $\vec{\theta^*}$ are the \basemodel{} parameters, $\vec{\theta}$ are the updated parameters and $\vec{\Omega}$
is the importance of each parameter.
$(\cdot)^{\odot 2}$ is the element-wise square (Hadamard square).
Intuitively, this loss penalizes changing the network parameters away from their initial values, where the penalty is higher for important parameters.
We compute $\vec{\Omega}$ using Memory-Aware Synapses~(MAS)~\cite{aljundi18eccv}%
, which estimates importance based on how much changes to the parameters affect the prediction of the model.

\para{Combined loss.}
Our full method uses a linear combination of the above losses:
\begin{equation}
\begin{aligned}
\Ladapt(\bx, \bp, \bc; \btheta) = \lambda\GCE(\bx, \bc; \btheta) 
 + (1-\lambda) \GCE(\bx, \bp; \btheta)  
 + \gamma \MAS(\btheta)
\end{aligned}
\label{eq:adapt-loss}
\end{equation}
where $\lambda$ balances the importance of the user corrections \vs the predicted mask, and $\gamma$ defines the strength of parameter regularization.
Next, we introduce \emph{\imageSGD{}} and \emph{\datasetSGD{}}, which both minimize Eq.~\eqref{eq:adapt-loss}. Their difference lies in how the model parameters $\btheta$ are updated: individually for each object or over a sequence.

\subsubsection{Adapting to a single image.}
\label{sec:imageSGD}
We adapt the segmentation model to a particular \changed{object in an image} by training on the click corrections.
We start from the segmentation model with parameters $\vec{\theta^*}$ fit to the initial training set (Sec.~\ref{sec:base_network}).
Then we update them by running several gradient descent steps to minimize \changed{our combined loss Eq.}~\eqref{eq:adapt-loss} every time the user makes a correction
(Algo. in supp. material).
We 
choose the learning rate and the number of update steps such that the updated model adheres to
the user corrections. This effectively turns corrections into constraints.
This process results in a segmentation mask $\bp$, predicted using the updated parameters $\btheta$. 

Adapting the model to the current test image brings two core advantages.
First, it learns about the specific appearance of the object and background in the current image.
Hence corrections have a larger impact and can also improve the segmentation of distant image regions which have similar appearance.
The model can also adapt to low-level photometric properties of this image, such as overall illumination, blur, and noise, which results in better segmentation in general.
Second, our adaptation step makes the corrections effectively hard constraints, so the model will preserve the corrected labeling in later iterations too.

This adaptation is done for each object separately, and the updated $\btheta$
is discarded once an object is segmented.

\subsubsection{Adapting to an image sequence.}
\label{sec:datasetSGD}
Here we describe how to continuously adapt the segmentation model to a sequence of test images using an online algorithm.
Again, we start from the model parameters $\vec{\theta^*}$  fit to the initial training set (Sec.~\ref{sec:base_network}).
When the first test image arrives, we perform interactive segmentation using these initial parameters.
Then, after segmenting each image $I_t=(\bx_t,\bc_t)$, the model parameters are updated to $\btheta_{t+1}$ by doing a single gradient descent step to minimize Eq.~\eqref{eq:adapt-loss} for that image. \changed{Thereby we subsample the corrections in the guidance maps to avoid trivial solutions (predict the corrections given the corrections themselves, see supp. material).}
The updated model parameters are used to segment the next image $I_{t+1}$.

\changed{Through the method described above our model adapts to the whole test image sequence, but does so gradually, as objects are segmented in sequence.}
As a consequence, this process is fast, does not require storing a growing number of images, and can be used in a online setting.
In this fashion it can adapt to changing appearance properties, adapt to unseen classes, and specialize to one particular class. It can even adapt to radically different image domains as we demonstrate in Sec.~\ref{sec:out_domain_adaptation}.

\subsubsection{Combined adaptation.}
For a test image $I_t$, we segment the object using \imageSGD{}~(Algo. in supp. material).
After segmenting a test image, we gather all corrections provided for that image and apply a image sequence adaptation step to update the model parameters from $\btheta_t$ to $\btheta_{t+1}$. %
At the next image, the image adaptation process will thus start from parameters $\btheta_{t+1}$ better suited for the test sequence.
This combination allows to leverage the distinct advantages of the two types of adaptation.

\subsection{Simulating user corrections}\label{sec:simulating_users}
To train and test our method we rely on simulated user corrections, as is common practice~\cite{xu16cvpr,liew17iccv,benard17arxiv,mahadevan18bmvc,li18cvpr,jang19cvpr}.

\para{Test-time corrections.}
When interactively segmenting an object, the user clicks on a mistake in the predicted segmentation.
To simulate this we follow~\cite{xu16cvpr,benard17arxiv,mahadevan18bmvc}, which assume that the user clicks on the largest error region.
We obtain this error region by comparing the model predictions with the ground-truth and select its center pixel.

\para{Train-time corrections.}
Ideally one wants to train with the same user model that is used at test-time.
To make this computationally feasible, we train the model in two stages \changed{as in}~\cite{mahadevan18bmvc}.
First, we sample corrections using ground-truth segmentations \cite{benard17arxiv,jang19cvpr,li18cvpr,liew17iccv,xu16cvpr}.
Positive user corrections are sampled uniformly at random on the object.
Negative user corrections are sampled
according to three strategies:
(1) uniformly at random from pixels around the object,
(2) uniformly at random on other objects,
and (3) uniformly around the object.
We use these corrections to train the model until convergence.
Then, we continue training by iteratively sampling corrections \changed{following}~\cite{mahadevan18bmvc}.
For each image we keep a set of user corrections $\bc$. Given $\bc$ we predict a segmentation mask, simulate the next user correction (as done at test time), and add it to $\bc$.
Based on this additional correction, we predict a new segmentation mask and minimize the loss (Eq.~\eqref{eq:ce-loss}).
Initially, $\bc$ corresponds to the corrections simulated in the first stage, and over time more user corrections are added. 
As we want the model to work well even with few user corrections, we thus periodically reset $\bc$ to the initial clicks~\cite{mahadevan18bmvc}.

\section{Experiments}
\label{sec:experiments}
We extensively evaluate our \imageSGD{} and \datasetSGD{} methods on several standard datasets as well as on aerial and medical images. These correspond to increasingly challenging adaptation scenarios.

\para{Adaptation scenarios.}
We first consider \textit{distribution shift}, where the training and test image sets come from the same general domain, consumer photos, but differ in their image and object statistics (Sec.~\ref{sec:in_domain_adaptation}).
This includes differences in image complexity, object size distribution, and when the test set contains object classes absent during training.
Then, we consider a \textit{\classspecialisation{}} scenario, where a sequence of objects of a single class has to be iteratively segmented 
(Sec.~\ref{sec:class_adaptation}).
Finally we test how our method handles large \textit{domain changes} where the imaging modality changes between training and testing. We demonstrate this by going from consumer photos to aerial and medical images (Sec.~\ref{sec:out_domain_adaptation}).

\para{Model Details.}  We use a strong re-implementation of~\cite{mahadevan18bmvc} as our interactive segmentation model (Sec.~\ref{sec:base_network}). We pre-train its parameters on \pascal{}~\cite{pascal-voc-2012} augmented with SBD~\cite{hariharan11iccv} (10582 images with 24125 segmented instances of 20 object classes).
As a baseline, we use this model as in~\cite{mahadevan18bmvc},~\ie without updating its parameters at test time.
We call this the \emph{\fixedmodellong}. This baseline already achieves state-of-the-art results on the PASCAL VOC12 validation set, simply by increasing the encoder resolution compared to~\cite{mahadevan18bmvc}
(3.44 clicks).
This shows that using a fixed set of model parameters works well when the train and test distributions match. 
We evaluate our proposed method by adapting the parameters of that same model at test time using \emph{single image adaptation} (IA), \emph{image sequence adaptation} (SA), and their combination (IA + SA).

\para{Evaluation metrics.}
We use
two standard metrics~\cite{xu16cvpr,liew17iccv,benard17arxiv,mahadevan18bmvc,li18cvpr,benenson19cvpr,jang19cvpr}:
(1) \textbf{\iouAtK{}}, the average intersection-over-union between the \gt{} and predicted segmentation masks, given $k$ corrections per image,
and (2) \textbf{\clicksAtIOU{}}, the average number of corrections needed to reach an IoU of $q\%$ on every image (thresholded at 20 clicks).
We always report mean performance over 10 runs (standard deviation is negligible at $\approx0.01$ for \clicksAtIOU{}).

\para{Hyperparameter selection.}
We optimize the hyperparameters for both adaptation methods on a subset of the \ade{} dataset \cite{zhou17cvpr,zhou18ijcv}.
Hence, the hyperparameters are optimized for adapting from \pascal{} to~\ade{}, which is distinct from the distribution shifts and domain changes we evaluate on.

\para{Implementation Details} are provided in the supplementary material.

\begin{table}[t]
	\centering
	\setlength{\tabcolsep}{3pt}
    \caption{\textbf{Adapting to distribution shifts.} 
	Mean number of clicks required to attain a particular mIoU score on Berkeley, YouTube-VOS and COCO datasets (Lower is better).
    Both of our adaptive methods, \imageSGD{}~(\imageadaptationshort{}) and \datasetSGD{}~(\sequenceadaptationshort{}) improve over the model that keeps the weights \fixedmodel{} at test time.
}
	\label{tab:distribution_adaptation}
	\resizebox{0.90\linewidth}{!}{%
	\setlength{\tabcolsep}{6pt}	
	\begin{tabular}{|l|c|c|@{}C{1.4cm}@{}C{1.4cm}@{}C{1.4cm}|}
		\toprule
& \textbf{\berkeley{} } & \textbf{\youtubeVOS}& \multicolumn{3}{c|}{\textbf{COCO~\cite{lin14eccv}}} \\
& \multicolumn{1}{c|}{\cite{mcguinness10book}} & \multicolumn{1}{c|}{\cite{xu18arxiv}} & seen & unseen & unseen 6k\\
		\arrayrulecolor{lightgray}\hline\arrayrulecolor{black}
		Method   & \clicksAt{90}\ & \clicksAt{85} &  \multicolumn{3}{c|}{\clicksAt{85}} \\
		\hline
\uc \fixedmodellong{}~\cite{mahadevan18bmvc} & 5.4 & 7.9 & 10.0 & 11.9 & 13.2 \\ %
		\arrayrulecolor{lightgray}\hline\arrayrulecolor{black}
\imageadaptationshort{}  & 4.9& 7.0 & 9.1 & 10.7 & 10.6 \\
\sequenceadaptationshort{}  & 5.3 & 6.9  & 9.7 & 10.6 &  10.0\\
\combinedshort{} & \textbf{4.9} & \textbf{6.7}  & \textbf{9.1} & \textbf{9.9}    & \textbf{9.3} \\
        \arrayrulecolor{lightgray}\hline\arrayrulecolor{black}
        $\Delta$ over \fixedmodellong{} 
		& \color{green!40!black}{8.5\%} & \color{green!40!black}{15.2\%}  & \color{green!40!black}{9.0\%} & \color{green!40!black}{16.8\%}    &\color{green!40!black}{29.5\%} \\
   		\bottomrule
	\end{tabular}
	}
\end{table}

\subsec{Adapting to distribution shift}
\label{sec:in_domain_adaptation}
We test how well we can adapt the model which is trained on \pascal{} to other consumer photos datasets.

\para{Datasets.}
We test on: (1) \emph{\berkeley{}}~\cite{mcguinness10book}, 100 images with a single foreground object.
(2) \emph{\youtubeVOS{}}~\cite{xu18arxiv}, a large video object segmentation dataset.
We use the test set of the 2019 challenge, where we take the first frame with ground truth (1169 objects, downscaled to $855\times480$ maximal resolution).
(3) \emph{\coco{}}~\cite{lin14eccv}, a large segmentation dataset with 80 object classes.
20 of those overlap with the ones in the \pascal{} dataset and are thus \emph{seen} during training. The other 60 are \emph{unseen}.
We sample 10 objects
per class from the validation set and separately report results for seen (200 objects) and unseen classes (600 objects) as in~\cite{xu16cvpr,majumder19cvpr}.
We also study how \datasetSGD{} behaves on longer sequences of 100 objects for each unseen class (named {\em \cocounseenlarge{}}).

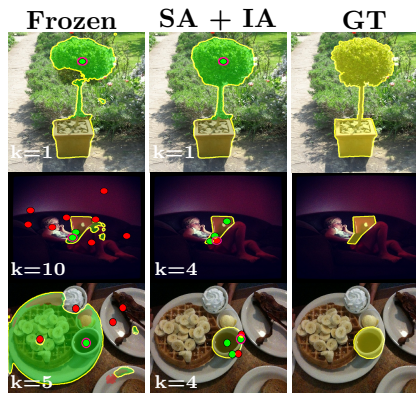
\begin{wrapfigure}{r}{0.48\linewidth}
    \begin{subfigure}[t]{\linewidth}
    \centering    
    \begin{overpic}[trim={0 0cm 0cm 0.5cm},clip,width=0.95\linewidth]{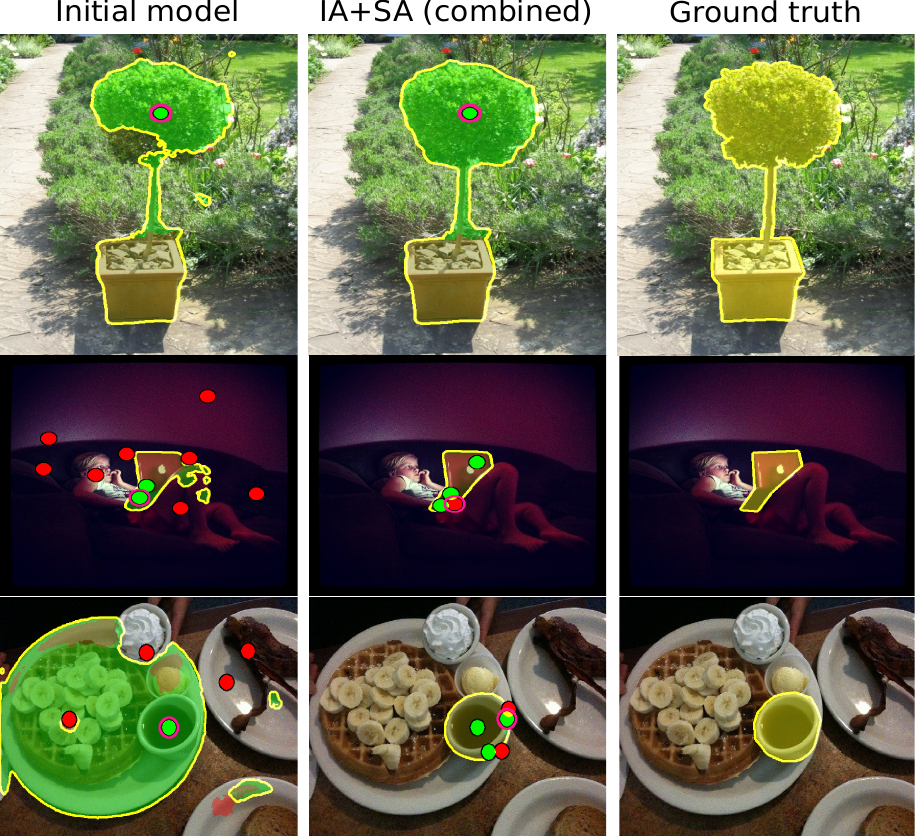}
    \put(150,900){\makebox(0,-20){\textbf{\uc \fixedmodel{}}}}
    \put(500,900){\makebox(0,-20){\textbf{SA + IA}}}
    \put(850,900){\makebox(0,-20){\textbf{GT}}}
    \put(55,580){\makebox(0,-10){\color{white}\textbf{\scriptsize k=1}}\color{black}}
    \put(395,580){\makebox(0,-10){\color{white}\textbf{\scriptsize k=1}}\color{black}}
    \put(70,300){\makebox(0,-10){\color{white}\textbf{\scriptsize k=10}}\color{black}}
    \put(400,300){\makebox(0,-10){\color{white}\textbf{\scriptsize k=4}}\color{black}}
    \put(55,30){\makebox(0,-10){\color{white}\textbf{\scriptsize k=5}}\color{black}}
    \put(400,30){\makebox(0,-10){\color{white}\textbf{\scriptsize k=4}}\color{black}}
    \label{fig:coco_unseen_hard}
    \end{overpic}
    \end{subfigure}    
    \caption{\textbf{Qualitative results} of the frozen and our combined adaptation model.
     Red circles are negative clicks and green ones are positive.
     Green and red areas respectively show the pixels that turned to FG/BG with the latest clicks.
     Our method produces accurate masks with fewer clicks \textbf{k}.
     }
    \label{fig:distribution_adaptations}
\end{wrapfigure}

\para{Results.}
We report our results in Tab.~\ref{tab:distribution_adaptation} and Fig.~\ref{fig:curve_collection_distribution}.
Both types of adaptation
improve performance on all tested datasets.
On the first few user corrections \textit{\imageSGD{}}~(\imageadaptationshort{}) performs similarly to the \fixedmodellong{} as it is initialized with the same parameters. But as more corrections are provided, it uses these more effectively to adapt its appearance model to a specific image.
Thus, it performs particularly well in the high-click regime,
which is most useful for objects that are challenging to segment (\eg due to low illumination, Fig.~\ref{fig:distribution_adaptations}),
or when~very~accurate~masks~are~desired.

\setlength{\intextsep}{2pt}%
\setlength{\columnsep}{10pt}%
During \textit{\datasetSGD{}}~(\sequenceadaptationshort{}), the model adapts to the test image distribution and thus learns to produce good segmentation masks given just a few clicks~(Fig.~\ref{fig:coco_unseen_6k_iou_at_k}).
As a result, \sequenceadaptationshort{} outperforms using a \fixedmodellong{} on all datasets with distribution shifts (Tab.~\ref{tab:distribution_adaptation}). By adapting from images to the video frames of \youtubeVOS{}, \sequenceadaptationshort{} reduces the clicks needed to reach 85\% IoU by \att{15\%}.
Importantly, we find that our method adapts fast, making a real difference after just a few images, and then keeps on improving even as the test sequence becomes thousands of images long (Fig.~\ref{fig:coco_unseen_6k_over_time}).
This translates to a large improvement given a fixed budget of 4 clicks per object: on the \cocounseenlarge{} split it achieves \att{69\%} IoU compared to the \att{57\%} of the \fixedmodellong{} (Fig.~\ref{fig:coco_unseen_6k_iou_at_k}).

Generally, the curves for \datasetSGD{} grow faster in the low click regime than the \imageSGD{} ones, but then exhibit stronger diminishing returns in the higher click regime (Fig.~\ref{fig:coco_unseen_6k_iou_at_k}).
Hence, combining the two compounds their advantages
leading to a method that considerably improves over the \fixedmodellong{} on the full range of number of corrections and sequence lengths (Fig.~\ref{fig:coco_unseen_6k_iou_at_k}).
Compared to the \fixedmodel{} model, our combined method significantly reduces the number of clicks needed to reach the target accuracy on all datasets: from a \att{9\%} reduction on Berkeley and COCO seen, to a \att{30\%} reduction on COCO unseen 6k.

\subsection{Adapting to a specific class}
\label{sec:class_adaptation}
When a user segments objects of a single class at test-time, \datasetSGD{} naturally specializes its appearance model to that class.
We evaluate this phenomenon on 4 \coco{} classes.
We form 4 test image sequences, each focusing on a single class, containing objects of varied appearance. The classes are selected based on how \datasetSGD{} performs compared to the \fixedmodellong{} in Sec.~\ref{sec:in_domain_adaptation}.
We selected the following classes, with increasing order of difficulty for \datasetSGD{}:
(1) donut (2540 objects)
(2) bench (3500)
(3) umbrella (3979)
and (4) bed (1450). 

\begin{figure}[t]
    \centering
    ~
   \begin{subfigure}[t]{0.3\textwidth}
    \includegraphics[width=1\linewidth]{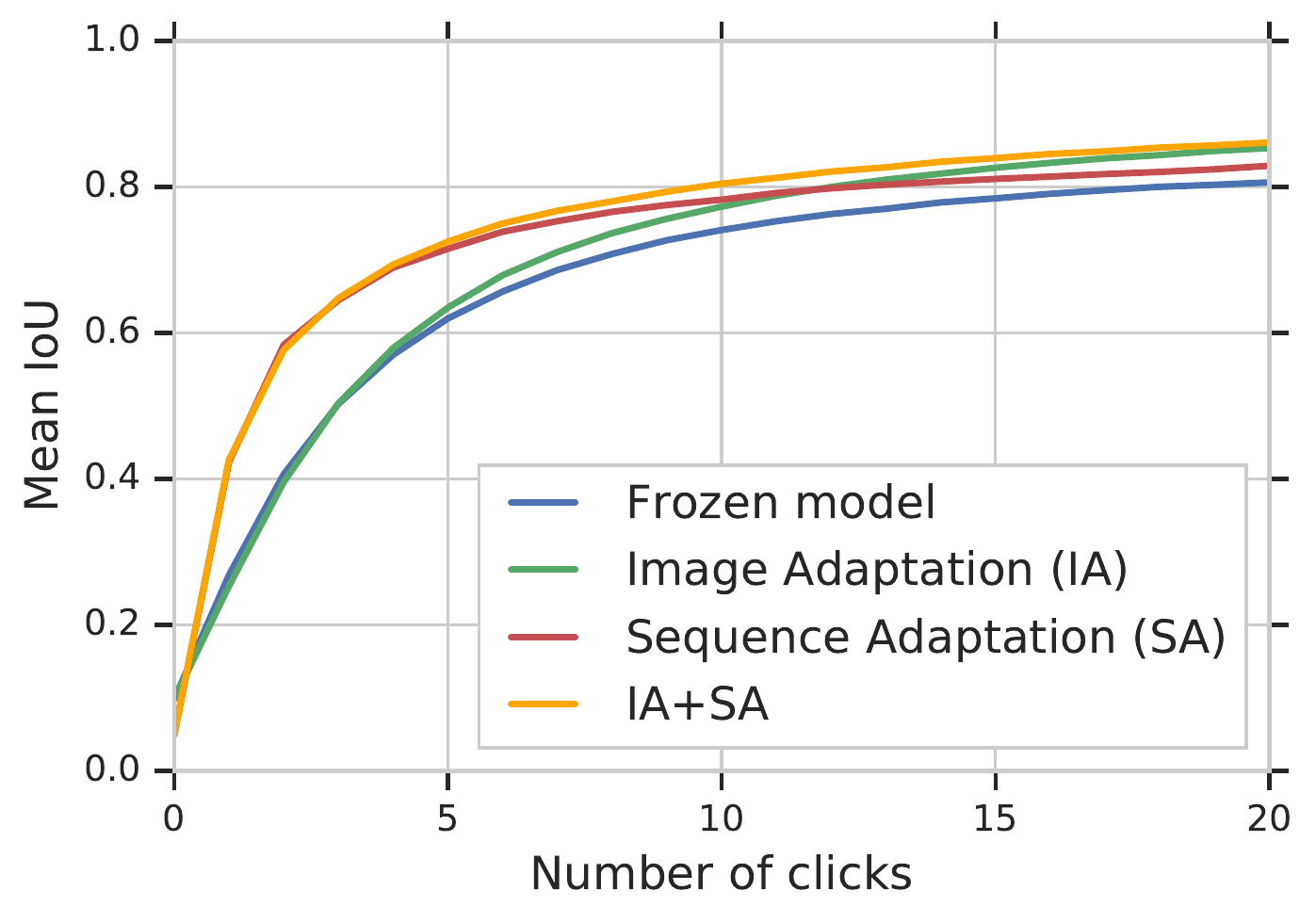}
    \caption{Mean \iouAtK{} for varying $k$ on \cocounseenlarge{}.
    Both forms of adaptation significantly improve over a frozen model.
    }
    \label{fig:coco_unseen_6k_iou_at_k}
    \end{subfigure}
    ~
    \begin{subfigure}[t]{0.3\textwidth}
    \includegraphics[width=1\linewidth]{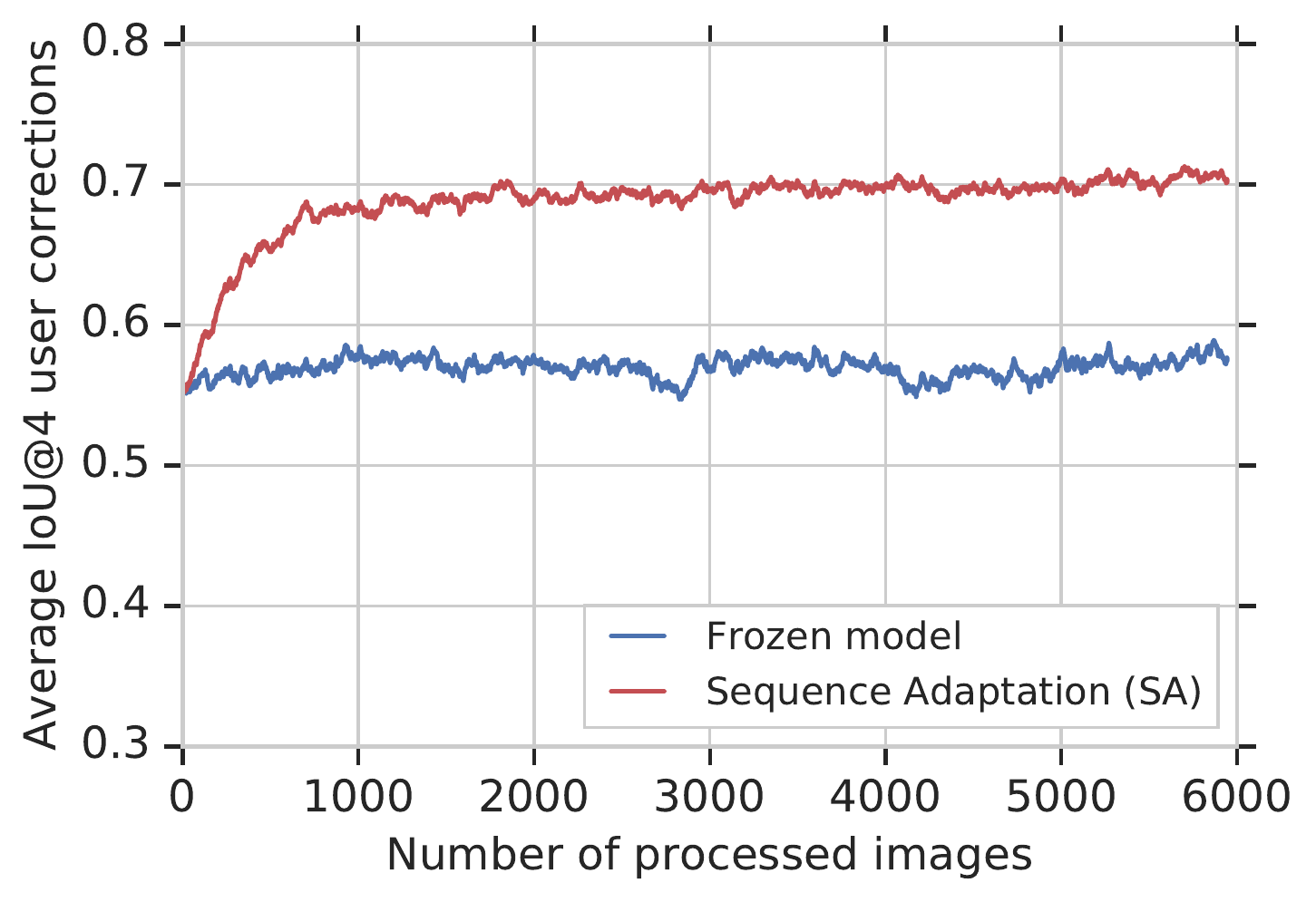}
    \caption{\iouAt{4} clicks as a function of the number of images processed.
    \sequenceadaptationshort{} quickly improves over the model with \fixedmodel{} weights.
    }
    \label{fig:coco_unseen_6k_over_time}
    \end{subfigure}
    ~    
    \begin{subfigure}[t]{0.3\textwidth}
    \includegraphics[width=1\linewidth]{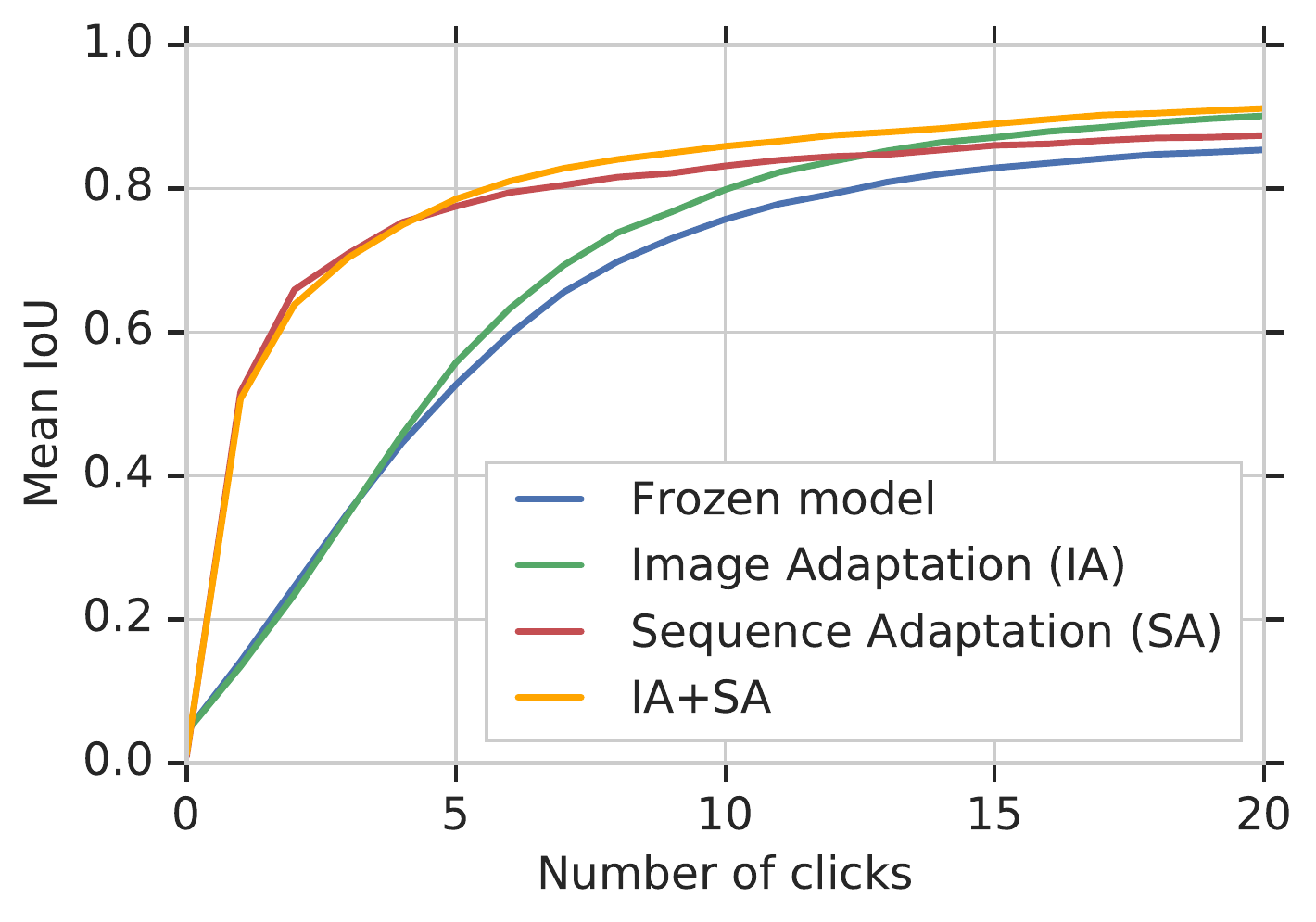}
    \caption{
    \iouAtK{} for varying $k$ when specializing to \textit{donuts}.	
    \sequenceadaptationshort{} offers large gains
    by learning a class specific appearance model.
    }
    \label{fig:coco_unseen_donut_clicks_plot}
    \end{subfigure}
    \caption{\textbf{Results for adapting to dist. shift (a,b) or a specific class (c).}
    } 
    \label{fig:curve_collection_distribution}
\end{figure}

\para{Results.}
Tab.~\ref{tab:class_adaptation_comparison}, Fig.~\ref{fig:coco_unseen_donut_clicks_plot}  present results.
The class specialization brought by our image sequence adaptation~(\sequenceadaptationshort{}) leads to good masks from very few clicks.
For example, on the donut class it reduces \clicksAt{85} by \att{39\%} compared to the \fixedmodellong{} and by \att{44\%} when combined with \imageSGD{} (Tab.~\ref{tab:class_adaptation_comparison}).
Given just 2 clicks, \sequenceadaptationshort{} reaches \att{66\%} \iou{} for that class, compared to \att{25\%} \iou{} for the \fixedmodellong{} (Fig.~\ref{fig:coco_unseen_donut_clicks_plot}). %
The results for the other classes follow a similar pattern,
showing that \datasetSGD{} learns an effective appearance model for a single class.

\begin{figure}[t!]   
    \begin{subfigure}[t]{0.50\textwidth}
    \includegraphics[width=0.49\linewidth]{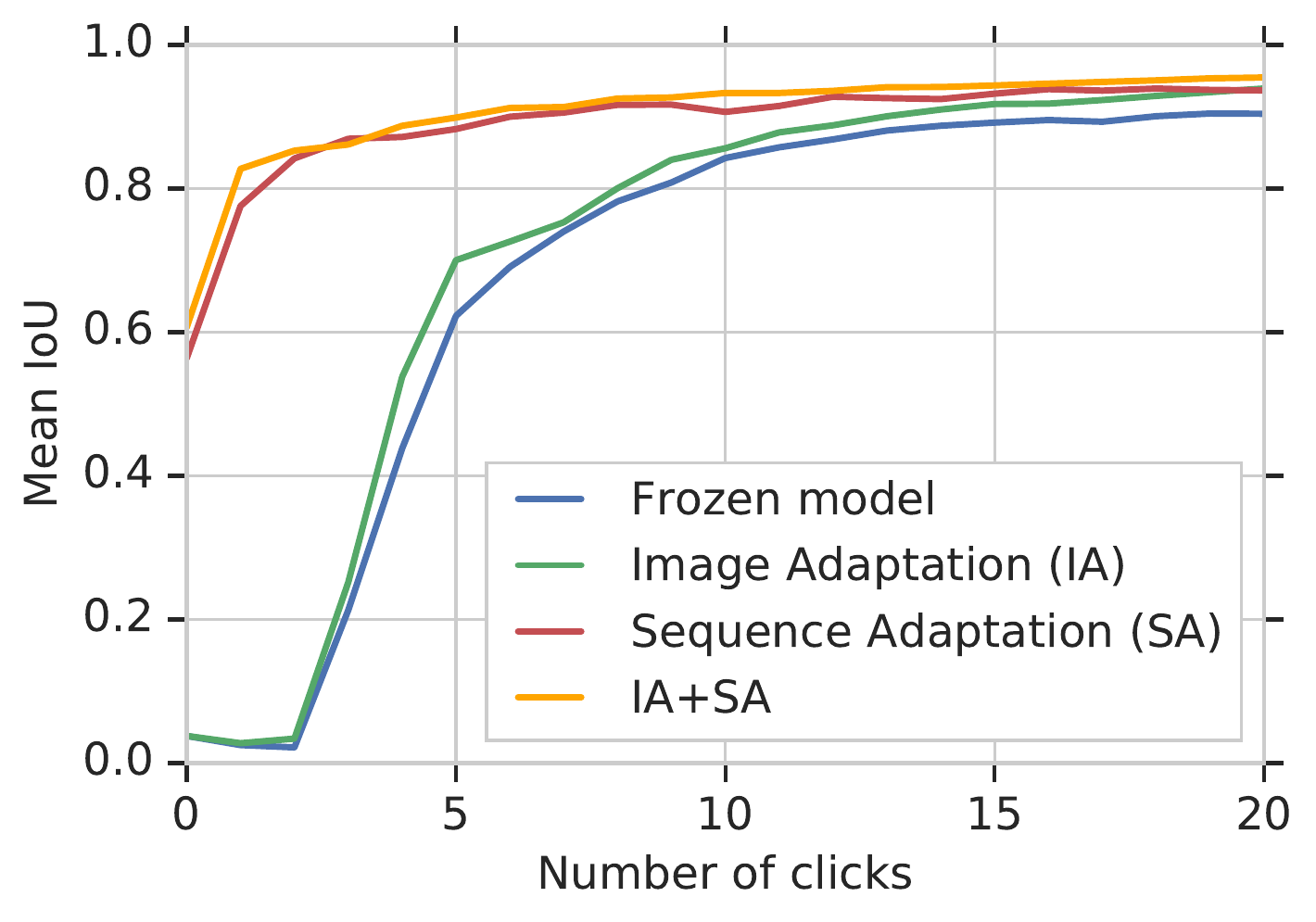}
    ~
    \includegraphics[width=0.47\linewidth]{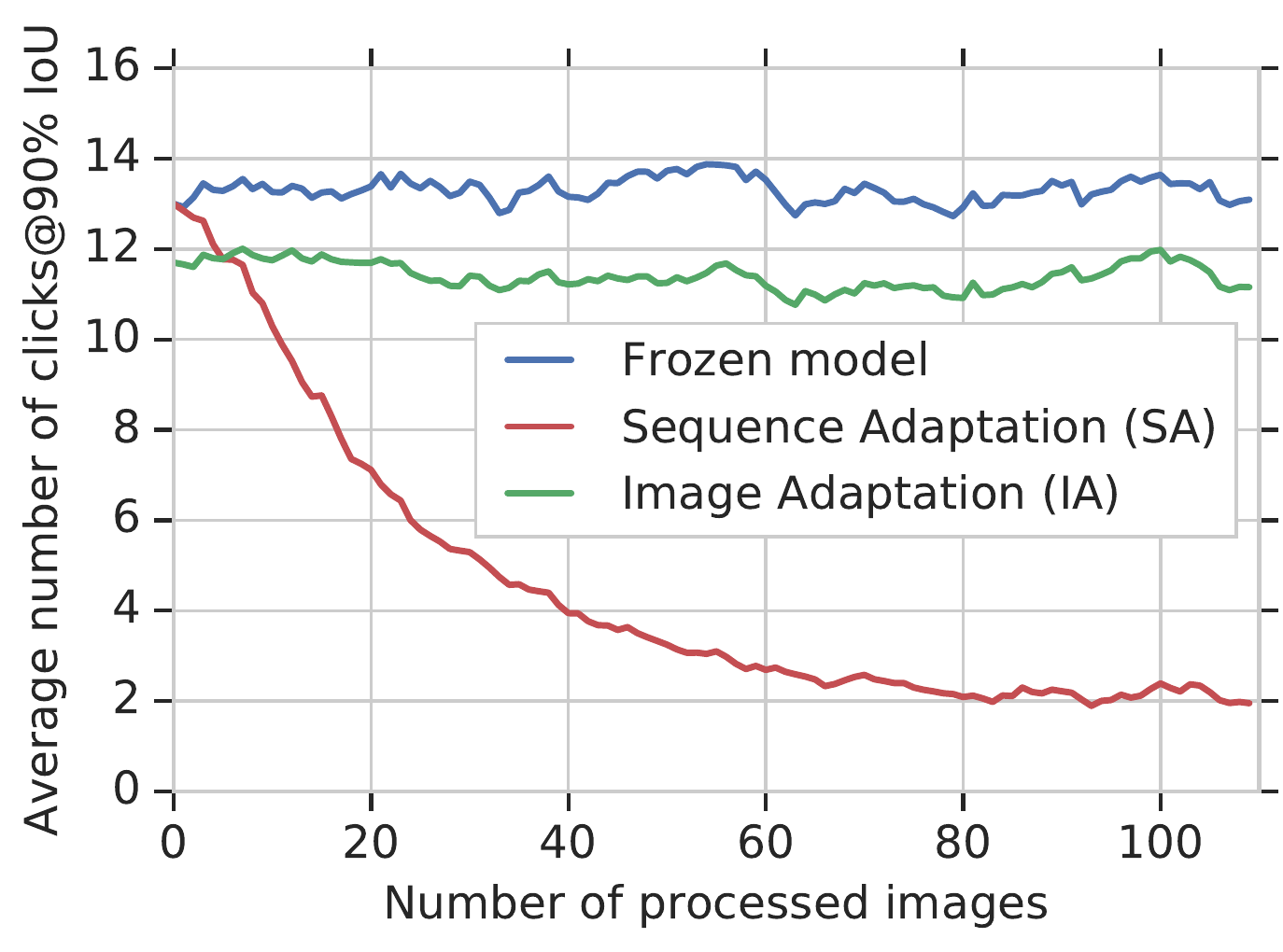}
    \caption{\medical{} dataset.
    }
    \label{fig:drions_sequence_adaptation}
    \end{subfigure}
    ~
    \begin{subfigure}[t]{0.50\textwidth}
    \includegraphics[width=0.49\linewidth]{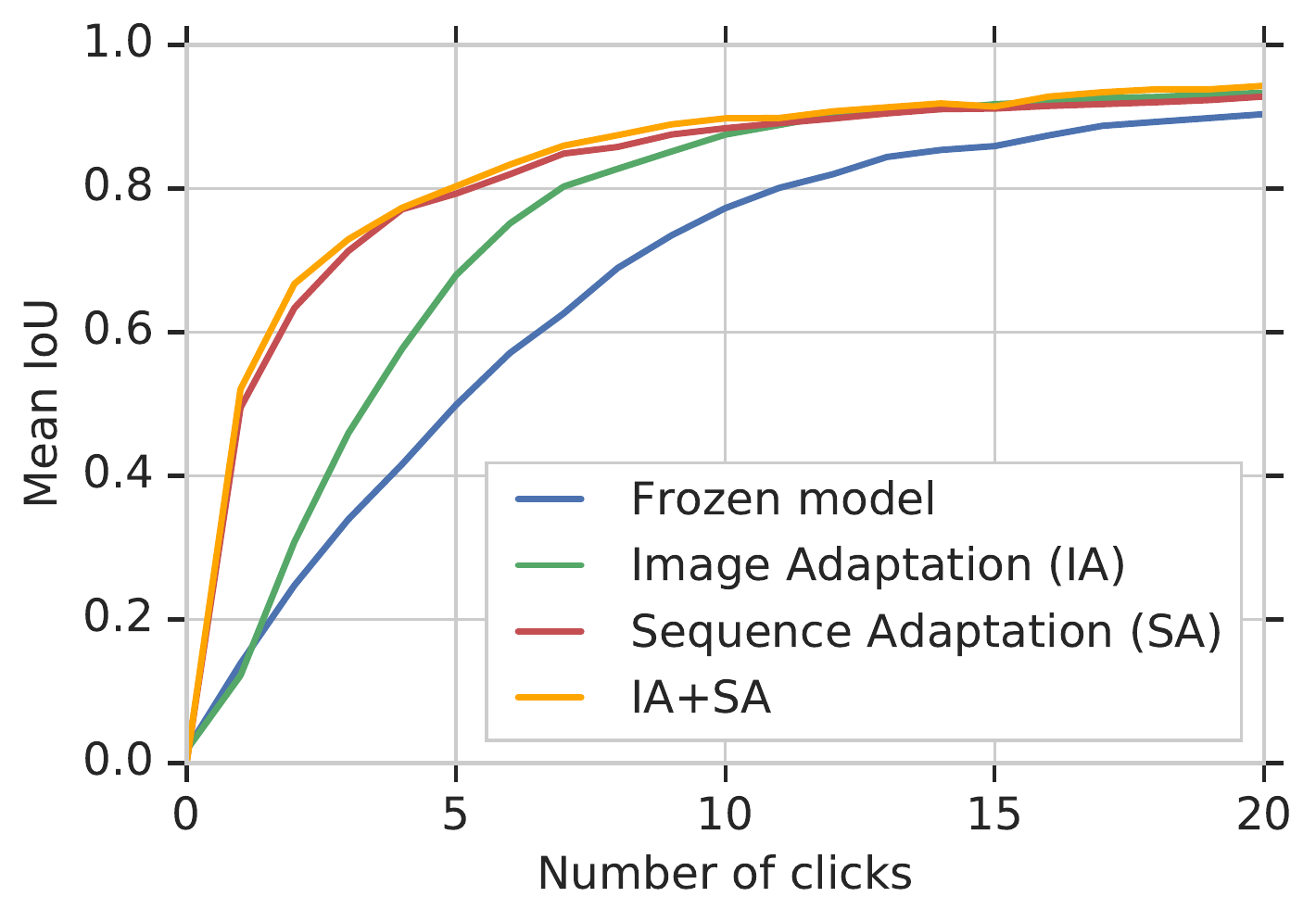}
    ~
    \includegraphics[width=0.47\linewidth]{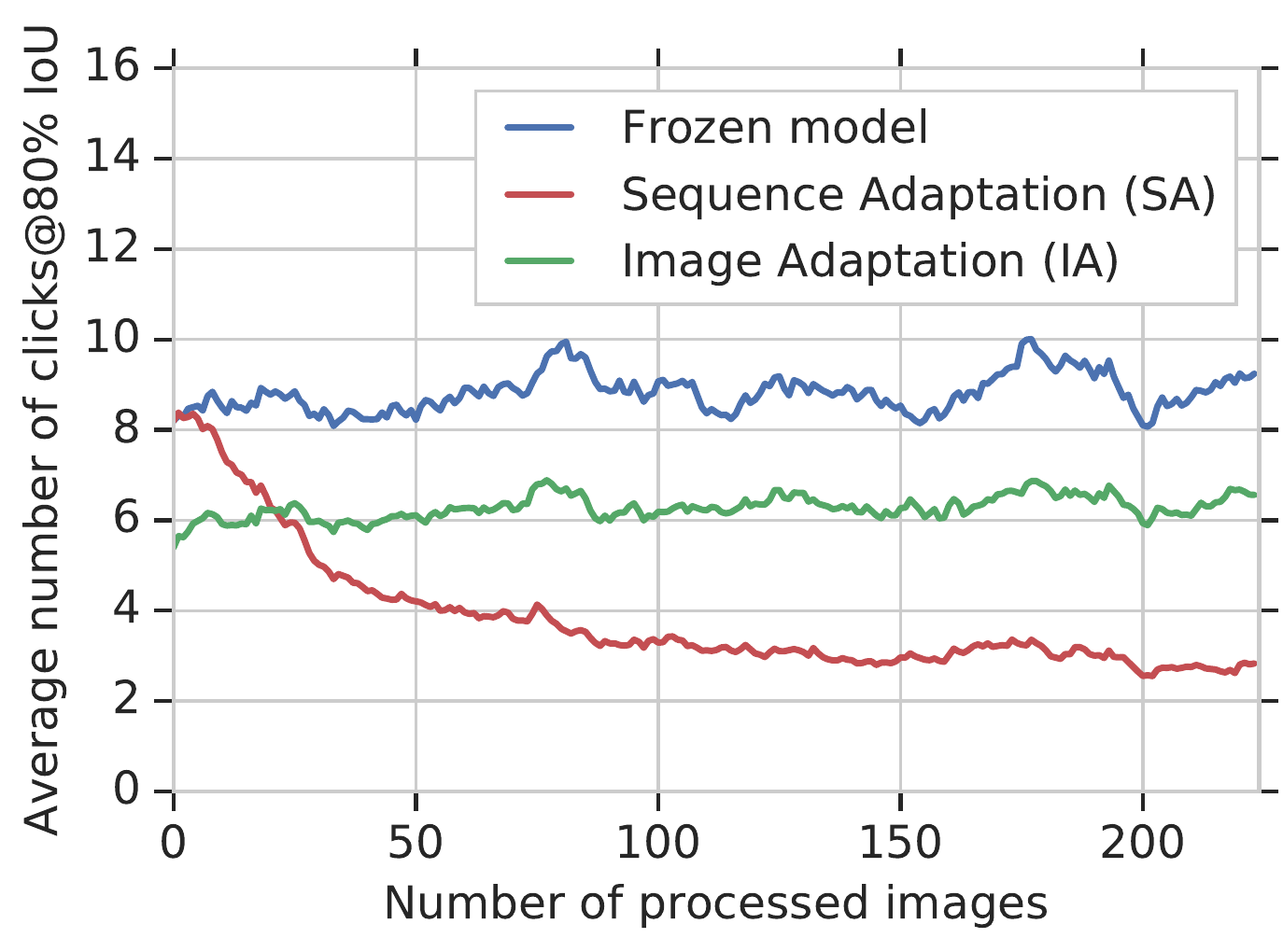}
    \caption{\rooftop{} dataset.
    }
    \label{fig:rooftop_improvement_smoothed}
    \end{subfigure}
    \caption{\textbf{Results for domain change.}
    For each dataset, we show the mean IoU at $k$ corrections (left in \ref{fig:drions_sequence_adaptation}, \ref{fig:rooftop_improvement_smoothed}) and the number of clicks to reach the target IoU as a function of the number of images processed (right in \ref{fig:drions_sequence_adaptation}, \ref{fig:rooftop_improvement_smoothed}).
    \uc \imageSGD{} provides a consistent improvement over the test sequences.
    Instead, \datasetSGD{} adapts its appearance model to the new domain gradually, improving with every image processed (right in \ref{fig:drions_sequence_adaptation}, \ref{fig:rooftop_improvement_smoothed}).
    }
    \label{fig:domain_adapt}

\end{figure}
\raggedbottom
\begin{table}[b]
	\centering
    \begin{minipage}{0.95\linewidth}
	\centering
	\caption{\textbf{Class specialization.} We test segmenting objects of only one specific class.
	Our adaptive methods outperforms the \fixedmodel{} model on all tested classes.
	Naturally, gains are larger for \datasetSGD{}, as it can adapt to the class over time.
}
	\label{tab:class_adaptation_comparison}
	\resizebox{0.8\linewidth}{!}{%
	\setlength{\tabcolsep}{6pt}
	\begin{tabular}{|p{3cm}r|c|c|c|}
		\toprule
        & \multicolumn{4}{c|}{\textbf{clicks @ 85\%~\iou{}}} \\
   & \multicolumn{1}{c|}{\textbf{Donut}} & \multicolumn{1}{c|}{\textbf{Bench}} & \multicolumn{1}{c|}{\textbf{Umbrella}} & \multicolumn{1}{c|}{\textbf{Bed}} \\
		\arrayrulecolor{lightgray}\hline\arrayrulecolor{black}
		\uc \fixedmodellong{}~\cite{mahadevan18bmvc} & 11.6 & 15.1 & 13.1 & 6.8 \\ %
		\arrayrulecolor{lightgray}\hline\arrayrulecolor{black}
		\imageadaptationshort{}~(Ours)          & 9.2 &   14.1 & 11.9 & 5.5 \\
		\sequenceadaptationshort{}~(Ours) & 7.1 & 14.0 & 11.1 & 5.5 \\
		\arrayrulecolor{lightgray}\hline\arrayrulecolor{black}
        \combinedshort{}~(Ours) &  \textbf{6.5} & \textbf{13.3} & \textbf{10.2} & \textbf{5.0} \\
        \arrayrulecolor{lightgray}\hline\arrayrulecolor{black}
        $\Delta$ over \fixedmodellong{} &  \color{green!40!black}{44.0\%} & \color{green!40!black}{11.9\%} & \color{green!40!black}{22.1\%} & \color{green!40!black}{26.5}\% \\
        \bottomrule

	\end{tabular}
	}
	\end{minipage}
\end{table}
\begin{table}[t]
	\centering
\begin{minipage}{0.90\linewidth}
	\centering
		\caption{\textbf{Domain change results}. We evaluate our model on 2 datasets that belong to different domains: aerial~(Rooftop) and medical~(DRIONS-DB). Both types of adaptation (\imageadaptationshort{} and \sequenceadaptationshort{})
		outperform the \fixedmodel{} model.%
	}
	\label{tab:domain_adaptation_results}
	\resizebox{0.80\linewidth}{!}{%
	\setlength{\tabcolsep}{6pt}
	\begin{tabular}{|p{3cm}cc|}
		\toprule
		& \textbf{DRIONS-DB~\cite{carmona08aim}} & \textbf{Rooftop~\cite{sun14eccv}} \\
		\textbf{Method}  & \clicksAt{90}~\iou{}  & \clicksAt{80}~\iou{}\\
		\arrayrulecolor{lightgray}\hline\arrayrulecolor{black}
		\uc \fixedmodellong{}~\cite{mahadevan18bmvc} & 13.3 & 8.9  \\ %
		\arrayrulecolor{lightgray}\hline\arrayrulecolor{black}
		\arrayrulecolor{lightgray}\hline\arrayrulecolor{black}
		\imageadaptationshort{}~(Ours) & 11.4 & 6.3\\
		\sequenceadaptationshort{}~(Ours) &  3.6 & 3.6 \\
		\arrayrulecolor{lightgray}\hline\arrayrulecolor{black}
		\combinedshort{}~(Ours) & \textbf{3.1}  & \textbf{3.6}  \\
		\arrayrulecolor{lightgray}\hline\arrayrulecolor{black}
        $\Delta$ over \fixedmodellong{} &  \color{green!40!black}{76.7\%} & \color{green!40!black}{59.6\%}  \\
		\bottomrule
	\end{tabular}
	}
	\end{minipage}
\end{table}
\subsec{Adapting to domain changes}
\label{sec:out_domain_adaptation}
We test our method's ability of adapting to domain changes by training on consumer photos (\pascal{}) and evaluating on aerial and medical imagery.  

\para{Datasets.} We explore two test datasets:
(1) \textit{\rooftop{}}~\cite{sun14eccv}, a dataset of 65 aerial images with segmented rooftops and 
(2) \textit{\medical}~\cite{carmona08aim}, a dataset of 110 retinal images with a segmentation of the optic disc of the eye fundus.
(we use the masks of the first expert). 
Importantly, the initial model parameters $\vec{\theta^*}$ were optimized for the \pascal{} dataset, which consists of consumer photos. Hence, we explore truly large domain changes here.

\para{Results.}
Both our forms of adaptation significantly improve over the \fixedmodellong{} (Tab.~\ref{tab:domain_adaptation_results}, Fig.~\ref{fig:domain_adapt}). 
\uc \imageSGD{} can only adapt to a limited extent, as it independently adapts to each object instance, always starting from the same initial model parameters $\vec{\theta^*}$.
Nonetheless, it offers a significant improvement, reducing the number of clicks needed to reach the desired IoU by \att{14\%-29\%}.
\uc \datasetSGD{}~(\sequenceadaptationshort{}) shows extremely strong performance, as its adaptation effects accumulate over the duration of the test sequence.
It reduces the needed user input by \att{60\%} for the \rooftop{} dataset and by over \att{70\%} for \medical{}.
When combining the two types of adaptation, the reduction increases to \att{77\%} for the \medical{} dataset (Tab.~\ref{tab:domain_adaptation_results}).
Importantly, our method adapts fast: on \medical{}
\clicksAt{90}
drops quickly and converges to just \att{2} corrections, as the length of the test sequence increases (Fig.~\ref{fig:drions_sequence_adaptation}). 
In contrast, the \fixedmodellong{} performs poorly on both datasets.
On the \rooftop{} dataset, it needs even more clicks than there are points in the ground truth polygons (\att{8.9} \vs \att{5.1}). This shows that even a state-of-the-art model like~\cite{mahadevan18bmvc} fails to generalize to truly different domains and highlights the importance of adaptation.

To summarize: We show that our method can bridge large domain changes spanning varied datasets and sequence lengths.
With just a single gradient descent step per image, our \datasetSGD{} successfully addresses a major shortcoming of neural networks, for the case of interactive segmentation: Their poor generalization to changing distributions~\cite{recht18arxiv,alcorn19cvpr}.
\subsec{Comparison to Previous Methods}
\label{sec:stoa_experiments_comparison}
\begin{table*}[t]
	\centering
	\caption{
The focus of our work is handling distribution shifts and domain changes between training and testing (Tab.~\ref{tab:distribution_adaptation},~\ref{tab:class_adaptation_comparison} \& \ref{tab:domain_adaptation_results}).
For completeness, we also compare our method against existing methods on standard datasets, where the distribution mismatch between training and testing is small.
At the time of initially releasing our work~\cite{kontogianni19arxiv}, our method outperformed all previous state-of-the-art models on all datasets. %
Later, %
{F-BRS}~\cite{sofiiuk20cvpr} (CVPR 2020) achieved even  better results.
}
	\label{tab:stoa_comparison}
	\resizebox{0.8\linewidth}{!}{%
	\setlength{\tabcolsep}{2.5pt}
	\begin{tabular}{|p{5.2cm}|c|c|c|c|ccc|}
		\toprule
		& \textbf{VOC12~\cite{pascal-voc-2012}}  & \textbf{GrabCut~\cite{Rother04-tdfixed}} & \textbf{Berkeley~\cite{mcguinness10book}}& \multicolumn{1}{c|}{\textbf{DAVIS~\cite{perazzi16cvpr}}}\\
& \multicolumn{1}{c|}{validation} & & & 10\% of frames\\
		\arrayrulecolor{lightgray}\hline\arrayrulecolor{black}
		\textbf{Method} & \clicksAt{85} & \clicksAt{90} & \clicksAt{90} & \clicksAt{85}\\
		\hline
		iFCN w/ GraphCut~\cite{xu16cvpr} & 6.88 & 6.04 & 8.65 &-\\
		RIS~\cite{liew17iccv} & 5.12 & 5.00 & 6.03 & - & \\		
		TSLFN~\cite{hu19nn} & 4.58 & 3.76 & 6.49&- & \\
		VOS-Wild~\cite{benard17arxiv} & 5.6\hspace{1.4mm} & 3.8\hspace{1.4mm} & - & - \\		
		ITIS~\cite{mahadevan18bmvc} & 3.80 & 5.60 & -  & -& \\
		CAG~\cite{majumder19cvpr} &  3.62 & 3.58 & 5.60&- & \\
		Latent Diversity~\cite{li18cvpr} & - & 4.79& -&   5.95 \\		
		BRS~\cite{jang19cvpr} & - & 3.60 & 5.08 & 5.58  \\ %
		\arrayrulecolor{lightgray}\hline\arrayrulecolor{black}
		F-BRS~\cite{sofiiuk20cvpr} (Concurrent Work) & - & \color{white!30!black}{2.72} & \color{white!30!black}{4.57} & \color{white!30!black}{5.04}  \\ %
		\hline
	    \arrayrulecolor{lightgray}\hline\arrayrulecolor{black}
		\combinedshort{} combined~(Ours) &  \textbf{3.18} & \textbf{3.07}& \textbf{4.94}& \textbf{5.16} \\
		\bottomrule
	\end{tabular}
	}
\end{table*}

While the main focus of our work is tackling challenging adaptation scenarios, we also compare our method
against state-of-the-art interactive segmentation methods on standard datasets.
These datasets are typically similar to \pascal{}, hence have a small distribution mismatch between training and testing.

\para{Datasets.}
(1) \berkeley{}, introduced in Sec.~\ref{sec:in_domain_adaptation}
(2) \textit{\grabcut{}}~\cite{Rother04-tdfixed}, 49 images with segmentation masks.
(3) \textit{\davis{}}~\cite{perazzi16cvpr}, 50 high-resolution videos out of which we sample 10\% of the frames uniformly at random as in~\cite{li18cvpr,jang19cvpr} (We note that the standard evaluation protocol of \davis{} favors adaptive methods, as the same objects appear repeatedly in the test sequence.) and 
(4) \textit{\pascal{} validation}, with 1449 images.

\para{Results.}
Tab.~\ref{tab:stoa_comparison} shows results. Our adaptation method achieves strong results:
At the time of initially releasing our work~\cite{kontogianni19arxiv}, it outperformed all previous state-of-the-art methods on all datasets (it was later overtaken by~\cite{sofiiuk20cvpr}).
It brings improvements even when the previous methods (which have \fixedmodel{} model parameters) already offers strong performance and need less than 4 clicks on average (\pascal{}, \grabcut{}). The improvement on \pascal{} further shows that our method helps even when the training and testing distributions match exactly (the \fixedmodellong{} needs \att{3.44} clicks).

Importantly, we find that our method outperforms~\cite{li18cvpr,jang19cvpr}, even though we use a standard segmentation backbone~\cite{chen18eccv} which predicts at $\frac{1}{4}$ of the input resolution. Instead~\cite{li18cvpr,jang19cvpr} propose specialized network architectures in order to predict at full image resolution, which is crucial for their good performance~\cite{jang19cvpr}.
We note that our adaptation method is orthogonal to these architectural optimizations and can be combined with them easily.

\subsec{Ablation Study}
\label{sec:ablation}
We ablate the benefit of treating corrections as training examples (on \cocounseenlarge{}).
For this, we selectively remove them from the loss (Eq.~\eqref{eq:adapt-loss}).
For \imageSGD{}, this leads to a parameter update that makes the model more confident in its current prediction, but this does not improve the segmentation masks. Instead, training on corrections improves \clicksAt{85} from \att{13.2} to \att{10.6}.
For \datasetSGD{}, switching off the corrections corresponds to treating the predicted mask as ground-truth and updating the model with it.
This approach implicitly contains corrections in the mask and thus improves ~\clicksAt{85} from \att{13.2} for the \fixedmodellong{} to \att{11.9}.
Explicitly using correction %
offers an additional gain of almost \att{2} clicks, down to \att{10}.
This shows that treating user corrections as training examples is key to our method:
They are necessary for \imageSGD{} and highly beneficial for \datasetSGD{}.

\subsection{Adaptation speed}
\begin{wrapfigure}{r}{0.38\linewidth}
    \centering
    \includegraphics[trim={0cm 0.35cm 0cm 0.2cm},clip,width=1\linewidth]{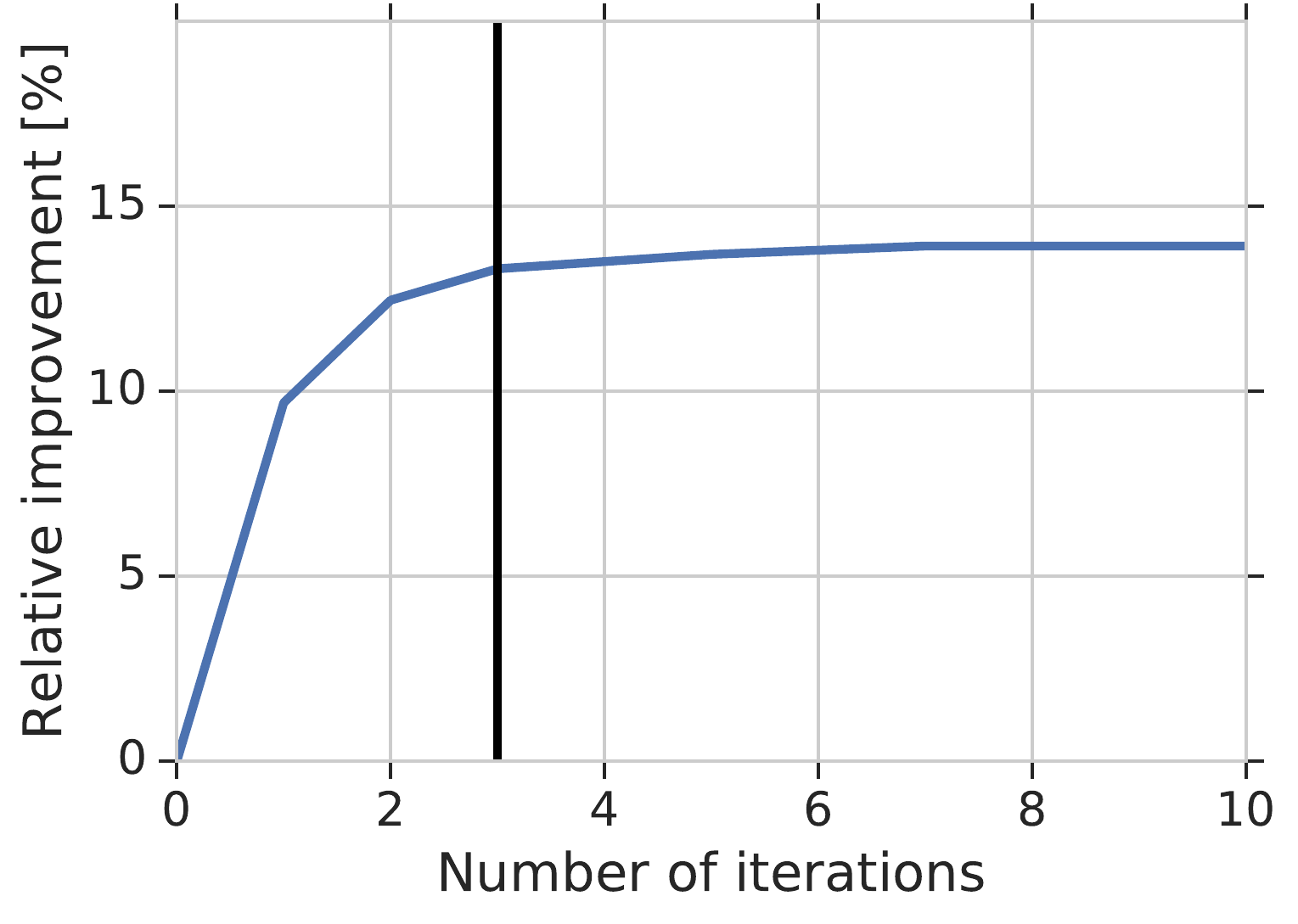}
    \caption{Iterations \vs relative improvement over a \fixedmodellong{} (mean over all datasets).}
    \label{fig:adaptation_iterations}
\end{wrapfigure}
While our method updates the parameters at test time, it remains fast enough for interactive usage.
For the model used throughout our paper %
a parameter update step takes 0.16\,s (Nvidia V100 GPU, mixed-precision training, \berkeley{} dataset).
\uc\datasetSGD{} only needs a single update step, done \emph{after} an object is segmented (Sec.~\ref{sec:datasetSGD}).
Thus, the adaptation overhead is negligible here.
For single image adaptation we used 10 update steps, 
for a total time of 1.6\,s.
We chose this number of steps based on hyperparameter search (see supp. material).
In practice, fewer update steps can be used to increase speed,
as they quickly show diminishing returns (Fig.~\ref{fig:adaptation_iterations}). 
We recommend to use 3 update steps, reducing adaptation time to $0.5$\,s, with a negligible effect on the number of corrections required (average difference of less than \att{1\%}, over all datasets).
\newline
\hspace{\parindent}
To increase speed further, the following optimizations are possible:
(1) Using a faster backbone,~\eg with a ResNet-50~\cite{he15arxiv}, the time for an update step reduces to 0.06\,s;
(2) Using faster accelerators such as Google Cloud TPUs;
(3) Employing a fixed feature extractor and only updating a light-weight segmentation head~\cite{li18cvpr}.

\section{Conclusion}
\label{sec:conclusion}
We propose to treat user corrections as sparse training examples and introduce a novel method that capitalizes on that idea to update the model parameters on-the-fly at test time.
Our extensive evaluation on 8 datasets shows the benefits of our method.
When distribution shifts between training and testing are small, our methods offers gains of \att{9\%-30\%}.
When specializing to a specific class, our gains are \att{12\%-44\%}. 
For large domain changes, where the imaging modality changes between training and testing, it reduces the required number of user corrections by \att{60\%} and \att{77\%}.

\para{Acknowledgement.}
We thank Rodrigo Benenson, Jordi Pont-Tuset, Thomas Mensink and Bastian Leibe for their inputs on this work.

\bibliographystyle{splncs04}
\bibliography{shortstrings,loco}

\iftoggle{appendsupplementary}{%
\pagebreak
\appendix
\section{Implementation Details}

\subsection{Subsampling for \datasetSGD{}}
We subsample the corrections in the guidance maps to avoid trivial solutions.
If all corrections were used equally in the loss and guidance maps, the model could eventually degrade to predict the corrections given the corrections themselves. Specifically it could degrade to only use the information in the guidance maps as its prediction, without relying on image appearance.
We avoid this by subsampling the clicks given to the network as guidance, but using all clicks to compute the adaptation loss (Eq. (4)). This forces the network to rely on appearance for propagating the corrections to the rest of the image, where the loss is sparsely evaluated at the pixel locations which were corrected.

\subsection{Robustness to image order}
Since our \textit{\datasetSGD{}}~(\sequenceadaptationshort{}) and \textit{combined adaptation}~(\combinedshort{})  methods are processing images sequentially, we tested our method's sensitivity to the image order. We repeated all our experiments 10 times by randomising the image order and computed the variance of the results. %
We found the variance to be minimal (\att{$\leq 0.01$} standard deviation) verifying that our adaptation methods are not sensitive to the order in which the images are processed. Hence, we only report averages to improve readability.

\subsection{Adaptation parameters}
Our adaptation methods use Adam \changed{optimizer}~\cite{kingma15iclr} with learning rate of $10^{-6}$ and batch size $1$.
For \imageSGD{} we do 10 SGD steps and regularize with $\lambda=1$ and $\gamma=1$. %
For \datasetSGD{} we do 1 SGD step and use $\lambda=0.5$  and $\gamma=2$.
For the \medical{} dataset we use a  learning rate of $10^{-5}$.

\subsection{\changed{Model details}}

We use DeeplabV3+~\cite{chen18eccv} with Xception-65~\cite{chollet17cvpr} as our backbone architecture (pre-trained on ImageNet~\cite{deng09cvpr} and \pascal{}~\cite{pascal-voc-2012}).
We extend this model with 2 extra channels for the guidance maps and train it for interactive segmentation model using SGD with momentum and an initial learning rate $0.0002$ with polynomial decay.
We use a batch size of~$2$, and atrous rates $\{12, 24, 36\}$. We use in input image resolution of $513\times513$ and an output stride of $8$ for the encoder and $4$ for the decoder, respectively.
For generating corrections, we sample at most $5$ foreground and $5$ background corrections for stage one of training (see Sec. 3.3 in the main paper). Corrections are encoded with disks of radius \att{3}.

\section{Comparison on the COCO dataset}
We have showed that all our adaptation methods are exhibiting substantial improvement compared to the \fixedmodellong{} in many datasets including the \coco{} dataset.
The improvement is especially large on the unseen classes of \coco~($16.8\%$ improvement, Table 1 in the main paper) and on adaptation to a particular unseen class ($44\%$ improvement for the \textit{donut} class, Table 2 in the main paper), two cases where adaptation is particularly useful.

While we outperform all previous methods on \pascal{}, \grabcut{}, \berkeley{} and \davis{}, some existing works report better \clicksAtIOU{} than us on COCO. \eg~\cite{li18cvpr} reports 7.86 compared to 9.69 for our method. We however note that these results are not directly comparable. 10 instances are sampled per class to form a test set and the selected instances have not been made available by previous works. But how the selection is done is crucial, as segmenting smaller objects is more challenging. If we ignore objects smaller than $80\times80$ pixels as in~\cite{benenson19cvpr}, for example, our \combinedshort{} improves from \att{9.9} to \att{5.4} (4.5 clicks less). %
Optimizing the architecture to better handle small objects is however not the focus of our work, as our adaptation methods work with any network architecture and can hence be combined with  architectural improvements easily.

\raggedbottom  
\section{\uc \imageSGD{} algorithm}

\begin{figure}[H]
\begingroup
\begin{minipage}{\linewidth}
\begin{algorithm}[H]
\scriptsize
\begin{algorithmic}[1]
\Function{SingleImageAdaptation}{input $\bx$, labels $\by$, target iou $\mathcal{J}^t$, initial parameters $\btheta^*$, learning rate $\lambda$, number of steps $k$}
  \State $\btheta \gets \btheta^*$ \Comment{Initialize adaptation model}
  \State $\bc \gets \vec{-1}$ \Comment{Start with no corrections}
  \For{$i \gets 0..20$}    \Comment{Iterate predicting and correcting}
      \State $\bp \gets \cnn(\bx; \btheta)$ \Comment{Predict mask}   
      \State $\mathcal{J} \gets $ \Call{IoU}{$\bp,\by$}  \Comment{Compute the IOU}
      \If{$\mathcal{J}\geq\mathcal{J}^t$} \Comment{Stop if mask has required IOU}
        	\State \Return {$(\bp, |\indicator[\bc \neq -1]|, \mathcal{J})$}
      \EndIf   
     \State $\bc \gets \bc$ $\cup $ \Call{GetCorrection}{$\bx,\bp,\bc$} \Comment{User input}
     \State $\bx \gets $ \Call{UpdateGuidance}{$\bx,\bc$}
     \For{$\mathrm{step} \gets 1..k$}     \Comment{Update model parameters}     
        \State $\btheta \gets \btheta - \lambda \frac{\deriv}{\deriv \btheta} \Ladapt(\bx, \bp, \bc; \btheta)$  
	 \EndFor       
	\EndFor
  	\State \Return {$(\bp, |\indicator[\bc \neq -1]|, \mathcal{J})$}		
\EndFunction
\end{algorithmic}
 \caption{\uc \imageSGD{}.}
 \label{algo:imageSGD}
\end{algorithm}
\end{minipage}
\endgroup
\end{figure}

\section{Number of corrections as proxy for segmentation time}
As is common practice~\cite{xu16cvpr,liew17iccv,benard17arxiv,mahadevan18bmvc,li18cvpr,jang19cvpr}, we rely on simulated user corrections to evaluate our method.
The number of corrections required to reach a certain segmentation quality serves as a proxy for the total time a user requires to segment an object. When less corrections are needed, segmenting an object is generally faster. But the time for making a correction might vary, as it comprises user interaction time (the time it takes for a user to make a correction) and computation time. We now contrast these two factors for our method.

Benenson \etal \cite{benenson19cvpr} performed a rigorous user study on interactive segmentation. There, they find that a user spends $\approx 3$ seconds per click correction (not including computation time). For the network used in our work, a single update step takes $0.16$ seconds (Sec. 4.6 in the main paper) and the forward pass takes ~0.04 seconds. Thus, single image adaptation requires a computation time of $\approx0.5s$ with 3 update steps. Image sequence adaptation is even faster as it only requires a single update step, which is done after an object is segmented.
Hence, given these timings, user interaction time is the dominating factor for the total segmentation time. Reducing the number of corrections required, as in our work, is therefore an effective way to save user time.

\section{Additional Qualitative Results}
We show additional results for our two adaptation methods compared to the \fixedmodellong{} in  Fig.~\ref{fig:additional_results}.
\begin{figure}
   \begin{subfigure}[t]{1.01\textwidth}
   \setlength{\abovecaptionskip}{-5pt plus 3pt minus 2pt} %
   \setlength{\belowcaptionskip}{20pt plus 3pt minus 2pt} %
     \centering
     \begin{overpic}[trim={0cm 0cm 0cm 1cm},clip,width=1\linewidth]{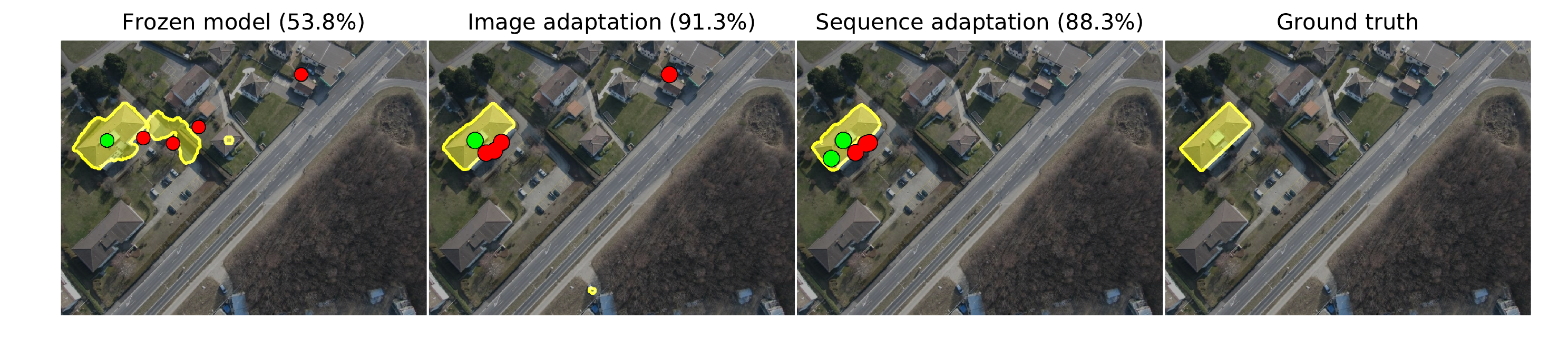}
       \put(150,220){\makebox(0,-20){\tiny Frozen Model~(53.8\%) } }
         \put(385,220){\makebox(0,-20){\tiny IA~(91.3\%)}}
        \put(620,220){\makebox(0,-20){\tiny SA~(88.3\%)}}
        \put(855,220){\makebox(0,-20){\tiny GT}}
        \end{overpic}
        \caption{\rooftop{} dataset~\cite{sun14eccv}}
        \label{fig:rooftop}

    \end{subfigure}  
    
   \begin{subfigure}[t]{1\textwidth}
    \setlength{\abovecaptionskip}{-5pt plus 3pt minus 2pt} %
    \setlength{\belowcaptionskip}{20pt plus 3pt minus 2pt} %
      \centering
       \begin{overpic}[trim={0cm 0cm 0cm 1cm},clip,width=1\linewidth]{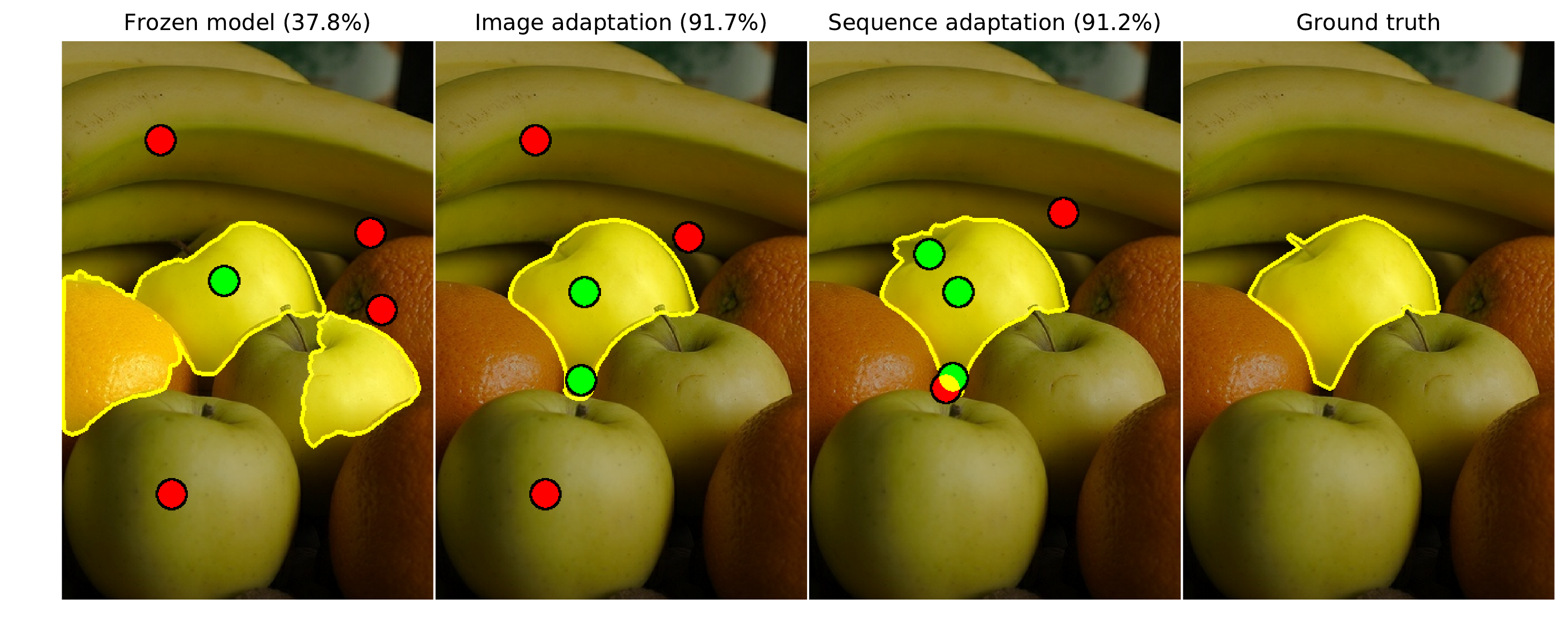}
        \put(150,400){\makebox(0,-20){\tiny Frozen Model~(37.8\%) } }
         \put(385,400){\makebox(0,-20){\tiny IA~(91.7\%)}}
        \put(620,400){\makebox(0,-20){\tiny SA~(91.2\%)}}
        \put(860,400){\makebox(0,-20){\tiny GT}}
    \end{overpic}
    \caption{\coco{} dataset~\cite{lin14eccv}}
        \label{fig:coco}
    \end{subfigure}
    
   \begin{subfigure}[t]{1\textwidth}
       \setlength{\abovecaptionskip}{-5pt plus 3pt minus 2pt} %
    \setlength{\belowcaptionskip}{10pt plus 3pt minus 2pt} %
    \begin{overpic}[trim={0cm 0cm 0cm 1cm},clip,width=1\linewidth]{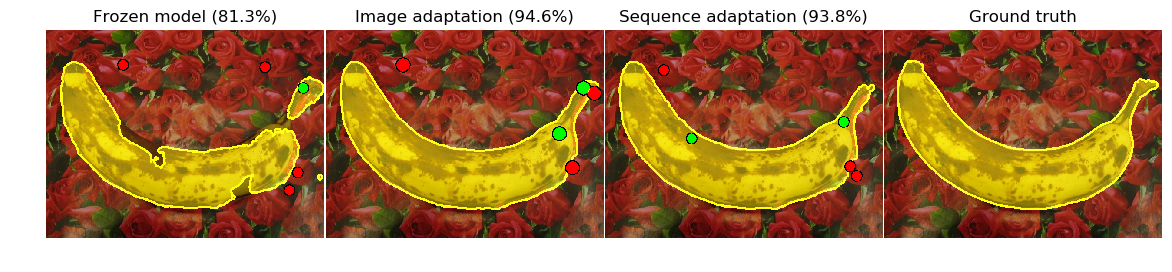}
           \put(150,220){\makebox(0,-20){\tiny Frozen Model~(81.3\%) } }
         \put(385,220){\makebox(0,-20){\tiny IA~(94.6\%)}}
        \put(620,220){\makebox(0,-20){\tiny SA~(93.8\%)}}
        \put(855,220){\makebox(0,-20){\tiny GT}}
    \end{overpic}
    \caption{GrabCut~\cite{Rother04-tdfixed}}
        \label{fig:grabcut}
    \end{subfigure}    
    
    \caption{\textbf{Additional results.}
    For each image we show results for our two adaptation methods (\imageadaptationshort{} and \sequenceadaptationshort)  and the \fixedmodellong{} for the same number of user corrections (\iouAt{5} is given in parenthesis).
    When the \fixedmodellong{} is applied to classes that are unseen during training, it sometimes produces segmentation masks that span multiple objects (\ref{fig:rooftop} \& \ref{fig:coco}) or do not respect object boundaries (\ref{fig:grabcut}).
\uc \imageSGD{}~(\imageadaptationshort) handles such cases much better, by adapting the model parameters to that specific object and its background. This allows it to correctly segment objects even when the foreground and background have similar appearance (\ref{fig:coco}).
\uc \datasetSGD{}~(\sequenceadaptationshort) optimizes the model parameters for the test sequence. This allows it to produce good masks from very few clicks, and additional clicks are only required close to the object to refine the exact boundary (\ref{fig:rooftop} \& \ref{fig:grabcut}).
    }
    \label{fig:additional_results}
\end{figure}

}{}

\end{document}